\documentclass{article}

     \PassOptionsToPackage{numbers, compress}{natbib}

\usepackage[main, final]{neurips_2026}

\usepackage[utf8]{inputenc} 
\usepackage[T1]{fontenc}    
\usepackage{hyperref}       
\usepackage{url}            
\usepackage{booktabs}       
\usepackage{amsfonts}       
\usepackage{nicefrac}       
\usepackage{microtype}      
\usepackage{xcolor}         
\usepackage{graphicx}       
\usepackage{amsmath}
\usepackage{tcolorbox}
\usepackage{subcaption}
\usepackage{multirow}
\usepackage{CJKutf8}

\title{ReflectMT: Internalizing Reflection for Efficient and High-Quality Machine Translation}

\author{%
  Kunquan Li$^{1,3}$ \thanks{Equal contribution.} \hspace{0.5em}\thanks{Work was done while Kunquan Li was an intern at WeChat AI, Tencent Inc.,China.} \quad Yingxue Zhang$^{2}$\footnotemark[1] \quad Zhibin Lan$^{1,3}$ \quad Fandong Meng$^{2}$\thanks{Corresponding author.} \quad \textbf{Jinsong Su}$^{1,3}$\footnotemark[3]
  \\
  $^{1}$School of Informatics, Xiamen University, China\\
  $^{2}$WeChat AI, Tencent Inc, China\\
  $^{3}$Key Laboratory of Digital Protection and Intelligent Processing of Intangible Cultural
 \\
 Heritage of Fujian and Taiwan (Xiamen University), Ministry of Culture and Tourism, China\\
  \texttt{\{likunquan,lanzhibin\}@stu.xmu.edu.cn, jssu@xmu.edu.cn} \\
  \texttt{\{yxuezhang,fandongmeng\}@tencent.com} \\
}

\begin{document}

\maketitle

\begin{abstract}
    Recent years have witnessed growing interest in applying Large Reasoning Models (LRMs) to Machine Translation (MT). While most approaches adopt a ``\textit{pre-thinking}'' paradigm and benefit from explicit reasoning trajectories, they suffer from substantial inference cost and latency. To address these limitations, we propose ReflectMT, a two-stage reflection internalization framework for machine translation that employs a ``\textit{post-thinking}'' paradigm. Our approach develops the model's ``\textit{translate–reflect–refine}'' capability through reinforcement learning. In the first stage, we cultivate the model's capacity for high-quality reflection and refinement, thereby enhancing its semantic comprehension and task-specific knowledge. In the second stage, we train the model to internalize the knowledge acquired during reflection. On WMT24, ReflectMT achieves a GPT-4o-based GRF score of 86.16, outperforming DeepSeek-R1 (84.03) by 2.13 points, while using approximately 92\% fewer output tokens. Independent automatic evaluation, a blinded human evaluation, and preliminary experiments on three additional translation directions provide further evidence for the framework's effectiveness.

\end{abstract}

\section{Introduction}
\label{intro}

In recent years, Large Reasoning Models (LRMs), such as OpenAI-o1~\citep{openai_o1_2024} and DeepSeek-R1~\citep{guo2025deepseek}, have demonstrated powerful capabilities in complex tasks such as mathematics, programming, and logical reasoning. These models typically leverage long Chain-of-Thought (CoT) processes, exploring different solution paths through reasoning, self-verification, and iterative refinement before generating the final answer.

Inspired by these advances, researchers have begun exploring the introduction of long CoT into Machine Translation (MT). We categorize these methods as the ``\textit{pre-thinking}'' paradigm, wherein the model performs explicit reasoning before generating the translation. For instance, Marco-o1~\citep{zhao2024marco} validated the effectiveness of long CoT in slang translation; \citet{wang2024drt} trained the DRT model through supervised fine-tuning for literary translation;  However, although the explicit reasoning process significantly enhances the model's ability to handle difficult translation tasks, the ``\textit{pre-thinking}'' paradigm requires the model to generate explicit and lengthy chains of thought during the inference stage, resulting in substantial computational overhead and latency. Although recent studies in general domains have explored efficient reasoning strategies---such as adaptive mode selection~\citep{zhang-etal-2025-adaptthink} and dynamic reasoning depth control~\citep{zhang2025adaptive}---they merely regulate the length or trigger of reasoning and overlook the multi-dimensional thinking aspects (e.g., semantic fidelity, cultural adaptation) that are central to translation, leaving the efficiency–quality trade-off in MT unresolved.

To address the above limitations, this paper proposes a novel ``\textit{post-thinking}'' algorithm for machine translation, named ReflectMT. Unlike the ``\textit{pre-thinking}'' paradigm, our core motivation is to leverage the reflection mechanism to enhance model capabilities during training, enabling the direct generation of high-quality translations during inference without incurring additional computational costs. The framework is divided into two phases, in the first stage (reflection capability establishment), we guide the model to learn the complete ``\textit{translate-reflect-refine}'' process through carefully designed reward functions. In the second stage (capability internalization and transfer), we guide the model to integrate the acquired reflection knowledge into the initial translation process by reinforcing the reward signal for direct translation. 

Specifically, we adopt a ``\textit{translate-reflect-refine}'' workflow during the training stage: the model first generates an initial translation, then initiates a structured reflection process to critically evaluate it, and finally generates a refined translation. Different from free-form reflection in general tasks, we design a multi-dimensional structured reflection process tailored to the specific needs of MT (covering error identification, ambiguity analysis, etc.) and construct a high-quality reflection dataset through multi-agent collaboration.

Furthermore, to achieve internalization of reflection capabilities, this algorithm mirrors the skill development of human translators: novices rely on explicit reflection and revision, whereas accumulated experience allows these processes to be internalized, enabling the direct generation of high-quality translations. This training paradigm achieves the transformation from ``\textit{explicit reflection}'' to ``\textit{implicit capability}'', enabling the model to directly output high-quality translations during inference.

Experimental results on multiple English-to-Chinese benchmarks show that ReflectMT substantially reduces inference-token consumption while maintaining competitive translation quality. On WMT24, ReflectMT achieves a GPT-4o-based GRF score of 86.16, compared with 84.03 for DeepSeek-R1, using 47.57 rather than 605.42 output tokens per sentence. Under the matched explicit-reflection setting, ReflectMT obtains a 0.24-point GRF advantage over Qwen3-8B with reflection (86.16 vs.\ 85.92), while reducing output tokens from 786.63 to 47.57 per sentence. Independent automatic evaluation, a blinded human evaluation, and preliminary experiments on three additional translation directions further support these findings.

The main contributions of this paper are summarized as follows:

\begin{itemize}
\item We introduce a ``\textit{post-thinking}'' reflection mechanism into neural machine translation, designing a multi-dimensional structured reflection process tailored to MT tasks that enables systematic evaluation and iterative improvement of translations.
\item We propose a reflection internalization training framework based on RL, which internalizes reflection knowledge into the model's direct generation capability, significantly improving translation quality while eliminating additional computational overhead during inference.
\item We construct a high-quality MT reflection dataset via multi-agent collaboration, providing a robust data foundation for cultivating the model's reflection capability. Extensive ablation studies on the dataset validate the effectiveness of our proposed framework. 
\end{itemize}

\section{Data Construction}
\label{subsec:2.2}

\begin{figure*}[t]
\centerline{\includegraphics[width=0.98\textwidth]{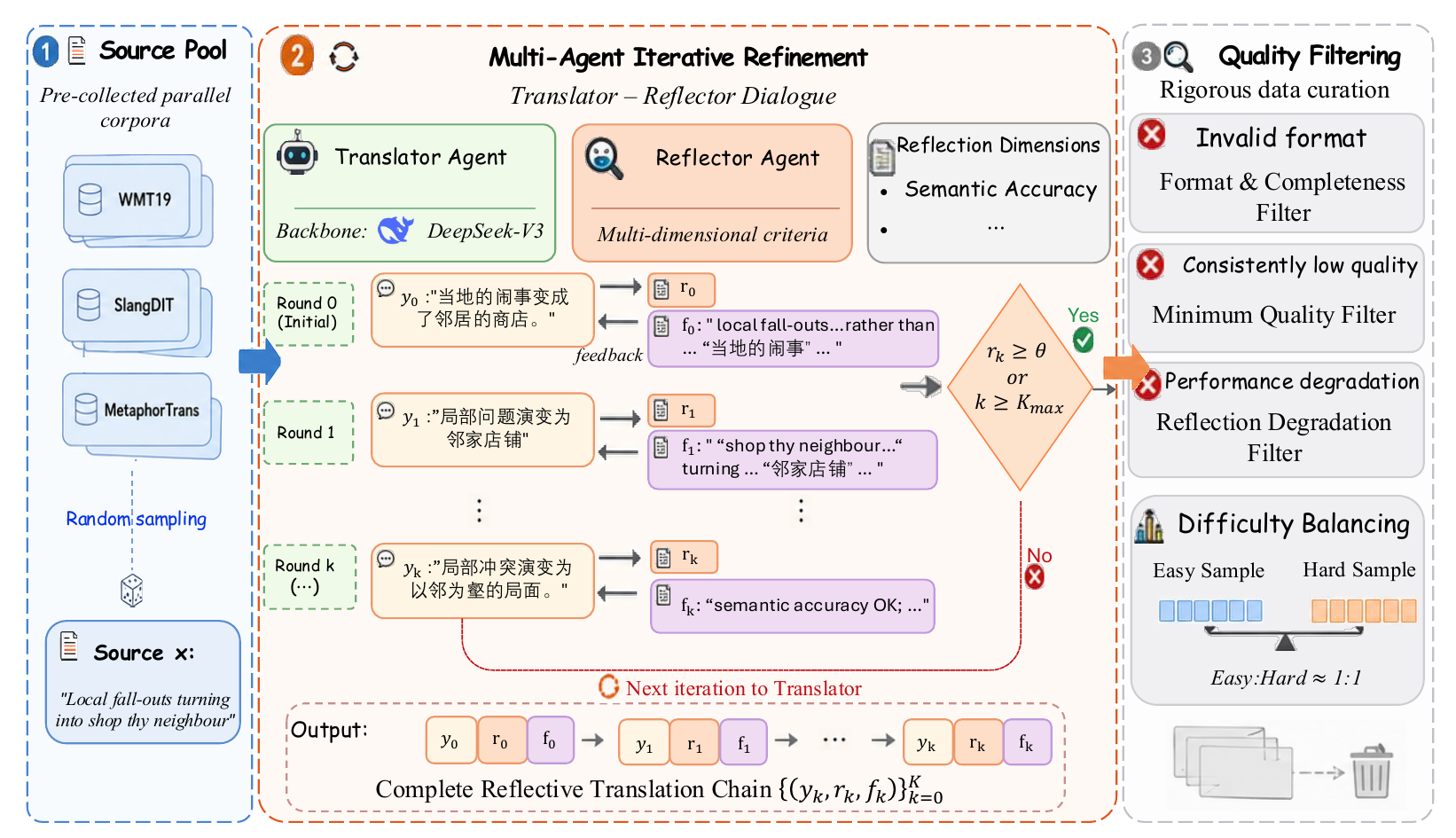}}
\caption{Overview of our reflective translation data construction. A Translator and a Reflector interact iteratively: the Translator produces a translation, the Reflector scores it and provides refinement suggestions, and this loop continues until the quality is satisfactory. The resulting data are then filtered to form the final reflective translation dataset.}
\label{fig:main}
\end{figure*}

High-quality reflective data is crucial for equipping models with structured reflection and refinement capabilities. However, existing machine translation datasets typically only contain parallel corpora and lack intermediate reasoning steps. To address this issue, we designed an iterative multi-agent collaboration system that automatically constructs datasets containing complete reflection chains through dialogues between the Translator and the Reflector.

As illustrated in Figure \ref{fig:main}, for each pre-collected sentence (denoted as \(x\)), we adopt a dual-agent framework for translation. The synthesis process is as follows:

\noindent
\textbf{(1)Initial Translation:} The Translator generates an initial translation of the source sentence \(x\), denoted as $y_0 = \text{Translator}(x)$.

\noindent
\textbf{(2)Reflective Evaluation:} The Reflector critically  evaluates \(y_0\) to provide a multi-dimensional assessment. This includes a quality score $r_0\!=\!\text{Reflector\_score}(x, y_0)\! \in \![0,100]$ based on predefined criteria (such as semantic accuracy, cultural adaptability, and fluency), alongside structured refinement suggestions $f_0 = \text{Reflector\_suggest}(x, y_0, r_0)$, that target specific lexical and structural deficiencies.

\noindent
\textbf{(3)Iterative Refinement Loop:} The agents collaboratively optimize the translation through a feedback loop. At each iteration $k$ ($k \geq 1$), the Translator generates an updated translation $y_k = \text{Translator}(x, y_{k-1}, f_{k-1}, r_{k-1})$.
The Reflector then returns a new score $r_k$ and updated suggestions $f_k$. This loop terminates when the score reaches a satisfactory threshold ($r_k \geq \theta$) or the maximum iteration limit is reached ($k \geq K_{\max}$). The sequence \(\{(y_k, r_k, f_k)\}_{k=0}^{K}\) forms a complete reflective translation chain.

To ensure data quality, we apply rigorous filtering to exclude: (1) samples with invalid formats; (2) samples with consistently low scores ($r_K < r_{low}$); and (3) samples exhibiting performance degradation after reflection ($r_K < r_0$).

Hyperparameter settings are detailed in Appendix \ref{app:Data}.

\section{Methodology}
\label{method}

\begin{figure*}[t]
\centerline{\includegraphics[width=1\textwidth]{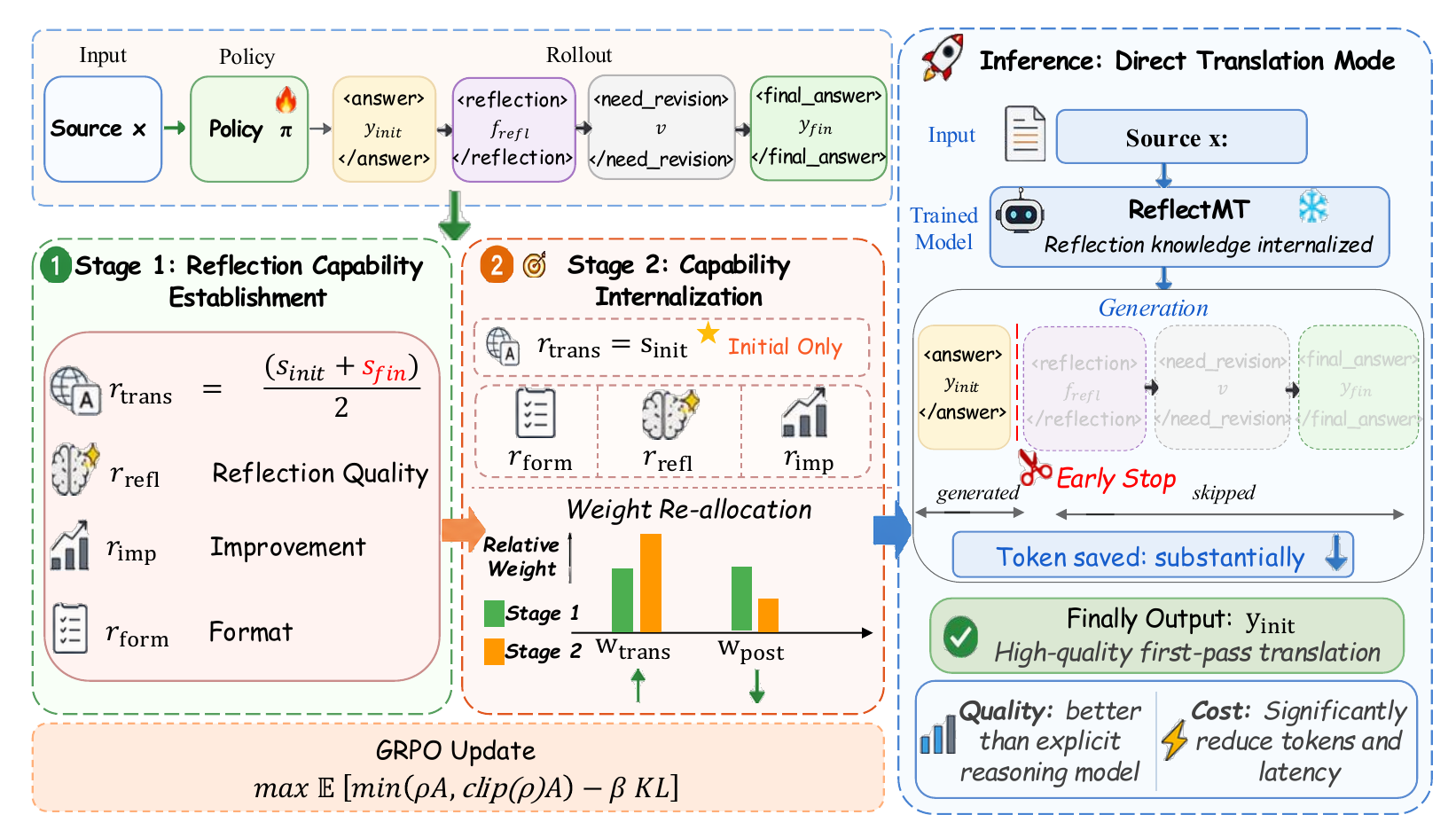}}
\caption{Training and inference paradigm of ReflectMT. During training, the model learns the complete ``\textit{translate--reflect--refine}'' process in two stages: Stage~1 establishes the reflection capability, and Stage~2 internalizes it into the initial translation. During inference, an early-stopping strategy is applied so that the model directly outputs the first-pass translations without explicit reflection.}
\label{fig:method}
\end{figure*}

\subsection{Overview}

Building on the reflective translation data from Section~\ref{subsec:2.2}, we present our \textit{Reflection Internalization} framework, which turns an explicit ``\textit{translate--reflect--refine}'' trajectory into a single-pass generation capability. As illustrated in Figure~\ref{fig:method}, the framework consists of (1) multi-dimensional reward modeling (Section~\ref{subsec:2.3.1}); and (2) a two-stage RL framework that first \emph{establishes} reflection (Section~\ref{subsec:2.3.2}) and then \emph{internalizes} it into the initial translation (Section~\ref{subsec:2.3.3}).



\subsection{Reward Modeling}
\label{subsec:2.3.1}

To implement our RL algorithm, we design four reward components. Given a source sentence \( x \), we require the model to generate outputs adhering to a strictly defined structural template: 
\begin{tcolorbox}
\fontsize{10pt}{11pt}\selectfont
<answer>\( y_{\text{init}} \)</answer>  <reflection>\( f_{\text{refl}} \)
</reflection>  <need\_revision>\( v_{\text{rev}} \)</need\_revision> 
  <final\_answer>\( y_{\text{fin}} \)</final\_answer>.
\end{tcolorbox}

We employ DeepSeek-V3~\cite{liu2024deepseek} as an LLM-as-a-Judge, denoted as \( \mathcal{J}_{\text{v3}}(\cdot) \), to score texts on a 0-1 scale.

\textbf{Format Reward (\( r_{\text{form}} \)).} We use regular expressions to verify whether the output strictly contains all required XML tags in the correct order: 
{
\begin{equation}
r_{\text{form}} = 
\begin{cases}
1 & \text{if format is correct} \\
0 & \text{otherwise}
\end{cases}
\ .
\end{equation}
}

\textbf{Reflection Quality Reward (\( r_{\text{refl}} \)).} We evaluate the reflection content \( f_{\text{refl}} \) based on the accuracy of problem identification and the actionability of the suggestions:
{
\begin{equation}
r_{\text{refl}} = {\mathcal{J}_{\text{v3}}(x, y_{\text{init}}, f_{\text{refl}})}.
\end{equation}
}

\textbf{Translation Quality Scores (\( s_{\text{init}} \) and \( s_{\text{fin}} \)).} We score the quality (e.g., semantic accuracy, fluency) of both the initial and final translations: \( s_{\text{init}} = \mathcal{J}_{\text{v3}}(x, y_{\text{init}}) \) and \( s_{\text{fin}} = \mathcal{J}_{\text{v3}}(x, y_{\text{fin}}) \). To facilitate a smooth transition from explicit reflection to implicit internalization, the formulation of the Translation Quality Reward (\( r_{\text{trans}} \)) dynamically adapts according to the training stage (detailed in Sections \ref{subsec:2.3.2} and \ref{subsec:2.3.3}).


\textbf{Reflection Improvement Reward (\( r_{\text{imp}} \)).}
To encourage meaningful refinement, we define the score difference
\( \Delta s = s_{\text{fin}} - s_{\text{init}} \).
The \texttt{<need\_revision>} tag encodes a binary decision
\( v_{\text{rev}} \in \{0,1\} \), where \(1\) indicates that the initial
translation requires revision. We apply the improvement reward only
when this decision is consistent with whether the final translation
actually modifies the initial translation:
\begin{equation}
r_{\text{imp}} =
\begin{cases}
\Delta s, & \text{if } v_{\text{rev}}
= \mathbf{1}[y_{\text{fin}} \neq y_{\text{init}}] \ \text{and} \ \Delta s > 0 , \\
0, & \text{otherwise}.
\end{cases}
\end{equation}

\textbf{Total Reward (\( R \)).} The total reward is a weighted sum of the aforementioned components:
{
\begin{equation}
\label{eq:4}
R \!=\!  r_{\text{form}} + w_{\text{trans}} r_{\text{trans}} + w_{\text{post}}( r_{\text{refl}} +  r_{\text{imp}}) \ ,
\end{equation}
}
where \( w_{\text{trans}}, w_{\text{post}}\) denote the corresponding weights assigned to each reward component. (see Appendix \ref{app:Details} for detailed hyperparameter settings).

\subsection{Stage 1: Reflection Capability Establishment}
\label{subsec:2.3.2}

In Stage 1, our objective is for the model to master the complete reflective translation task. We use Qwen2.5-7B-Instruct~\cite{yang2024qwen2} as the base model and perform a cold-start Supervised Fine-Tuning (SFT) via Low-Rank Adaptation (LoRA)~\cite{hu2021loralowrankadaptationlarge}, leveraging the data constructed in Section \ref{subsec:2.2}, enabling the model to generate the structured output format.

Subsequently, we apply the GRPO~\cite{shao2024deepseekmathpushinglimitsmathematical} algorithm. In this stage, we expect the model to produce a reasonable initial translation and a highly refined final translation. Therefore, the translation quality reward for Stage 1 is defined as the average of the initial and final translation scores:
{
\begin{equation}
r_{\text{trans}}^{\text{stage1}} = \frac{s_{\text{init}} + s_{\text{fin}}}{2} \ .
\end{equation}
}

Given a source sentence \( x \), GRPO samples \( n \) generations \(\{ g_1, \ldots, g_n \}\) from the policy \( \pi_{old} \). We compute the Stage 1 total reward \( r_i \) for each \( g_i \) using Equation \ref{eq:4}. GRPO optimizes the policy \( \pi_{\theta} \) by maximizing:
{
\begin{equation}
\begin{aligned}
\frac{1}{n} \sum_{i=1}^{n} \Bigg[ \min\Big( \rho_i A_i, \text{clip}(\rho_i, 1-\epsilon, 1+\epsilon) A_i \Big) - \beta \mathbb{D}_{\text{KL}}\bigl(\pi_{\theta} \big\| \pi_{\text{ref}}\bigr) \Bigg]
\end{aligned},
\end{equation}
}
\noindent where \( \rho_i = \frac{\pi_{\theta}(g_i | x)}{\pi_{old}(g_i | x)} \), \( \pi_{\text{ref}} \) is the cold-start SFT model, and \( \epsilon, \beta \) are hyperparameters. The advantage \( A_i \) is computed by normalizing the rewards:
{
\begin{equation}
A_i = \frac{r_{i} - {\operatorname{mean}(\{r_1, r_2, \cdots, r_n\})}}{{\operatorname{std}(\{r_1, r_2, \cdots, r_n\})}}.
\end{equation}
}
\subsection{Stage 2: Capability Internalization and Transfer}
\label{subsec:2.3.3}

In Stage 2, we shift the training focus to the quality of the initial translation. To force the model to internalize the reflection knowledge acquired in Stage 1 into its first-pass translations, we introduce two critical modifications to the reward mechanism.

First, the translation quality reward is solely determined by the initial translation score, disregarding the final translation:
{
\begin{equation}
r_{\text{trans}}^{\text{stage2}} = {s_{\text{init}}}.
\end{equation}
}

Second, we adjust the weight configuration of the total reward by significantly increasing the translation weight \( w_{\text{trans}}^{\text{new}} \) and correspondingly decreasing the improvement weight \( w_{\text{post}}^{\text{new}} \). However, the model is still required to output the full structured tags during training to maintain RL stability and prevent catastrophic forgetting of the reasoning format. This reward restructuring incentivizes the model to maximize the quality of the first-pass translations \( y_{\text{init}} \), thereby reducing its reliance on subsequent reflection steps.

\subsection{Inference Phase}
Owing to this internalization mechanism, the model directly produces high-quality first-pass translations. During inference, we externally terminate generation after the \texttt{</answer>} tag, retaining the complete translation while bypassing \texttt{<reflection>} and \texttt{<final\_answer>} to reduce computational overhead and latency. This is an inference-time policy rather than learned termination: without truncation, the model continues generating these fields. Internalization thus refers to translation knowledge, not stopping behavior.

\section{Experiments}
\label{exper}

\subsection{Experimental Setups}
\label{subsec:4.1}

\noindent \textbf{Data.}
\hspace{2mm}We constructed a reflective translation dataset specifically designed for English-to-Chinese machine translation, with the data volume shown in Table~\ref{tab:dataset}. Each sample comprises the source sentence, the initial translation, the reflective analysis, and the final translation. In addition, we used official test sets from WMT23\footnote{\url{https://www2.statmt.org/wmt23/translation-task.html}}, WMT24\footnote{\url{https://www2.statmt.org/wmt24/translation-task.html}}, and FLORES-200~\cite{nllbteam2022languageleftbehindscaling} to evaluate the model's generalization ability. To further probe generalization across translation directions, we additionally evaluated Chinese-to-English, English-to-Russian, and English-to-German translation, with results presented in Appendix~\ref{app:multilingual}. Comprehensive dataset statistics and qualitative examples are provided in Appendix~\ref{app:Data}.

\vspace{0.5ex}
\noindent \textbf{Metrics.}
Following \citet{wang2024drt}, we primarily evaluated the model's translation results using GRF (GPT Reference-Free), a GPT-based metric that does not require human references, with all GRF scores reported in this paper computed using GPT-4o. The evaluation prompt is detailed in Appendix \ref{app:Evaluation Prompt}. Additionally, we reported two widely used semantic-level evaluation metrics based on pre-trained models: COMETKIWI-XL(CK)~\cite{rei2023scalingcometkiwiunbabelist2023} and MetricX-24(MX24)~\cite{juraska2024metricx24googlesubmissionwmt}. In terms of efficiency, we evaluate using the number of tokens, calculated uniformly as ``\textit{average number of output tokens per sentence}''.

\vspace{0.5ex}
\noindent \textbf{Backbones.}
We use Qwen2.5-7B-Instruct as our base model. For detailed training information, please refer to Appendix \ref{app:Details}.

\subsection{Baselines}
\label{subsec:3.2}
To comprehensively evaluate the performance of our model, we selected three categories of baseline models, covering general large language models, large reasoning models, and specialized machine translation models.

\noindent
\textbf{General LLMs.} We use Llama3.1-8B~\cite{grattafiori2024llama}, Qwen2.5-7B, Qwen2.5-14B~\cite{yang2024qwen2}. Additionally, we compare with the closed-source commercial model GPT-4o~\cite{openai_o1_2024}, with results presented in Appendix~\ref{app:GPT}.

\noindent
\textbf{LRMs.} We use Marco-o1-7B~\cite{zhao2024marco}, QwQ-32B~\cite{team2024qwq}, DeepSeek-R1~\cite{guo2025deepseek}, and Qwen3-8B~\cite{yang2025qwen3} as baselines.

\noindent
\textbf{Specialized MT models.} Since existing efficient reasoning methods have not been explored in MT, we train two ``\textit{pre-thinking}'' baselines based on Qwen2.5-7B-Instruct: an \textit{Adapt} variant that learns to selectively engage in reasoning based on input complexity, and a \textit{Full} variant (i.e., a full-thinking mode; see Appendix \ref{app:Adapt}). We also include DRT-7B~\cite{wang2024drt}, DeepTrans-7B~\cite{wang2025deep}, and ExTrans-7B~\cite{wang2025extrans}, which follow a ``\textit{pre-thinking}'' paradigm.

To ensure a rigorous and equitable comparison, we additionally evaluated the baseline models under a ``\textit{translate-reflect-refine}'' setting, denoted as \textit{w/ refl}. Specifically, during inference, we apply an identical structured reflective prompt to the baselines, and we strictly use the resulting refined translation for evaluation. Notably, both our method and the baselines demonstrate perfect compliance with this format, consistently outputting the initial translation, reflective analysis, and final translation without exception.

\subsection{Main Result}
\label{subsec:3.3}

\begin{table*}[t]
\centering

\caption{Experimental results in English-to-Chinese translation. \textbf{Bold} and \underline{underlined} values denote the best and second-best scores, respectively. Superscripts $^{*}$, $^{**}$, and
$^{***}$ indicate statistically significant improvements over
DeepSeek-R1 with reflection under paired bootstrap tests at
$p<0.05$, $p<0.01$, and $p<0.001$, respectively.
Detailed results are provided in Appendix~\ref{app:Robustness Analysis},
Table~\ref{tab:bootstrap}.}

\resizebox{\textwidth}{!}
{
\begin{tabular}{lcccccccccccccccc}
\toprule[1pt]
\multirow{2}{*}{Model} & \multicolumn{4}{c}{Our} & \multicolumn{4}{c}{WMT23} & \multicolumn{4}{c}{WMT24} & \multicolumn{4}{c}{FLORES-200} \\
\cmidrule(lr){2-5} \cmidrule(lr){6-9} \cmidrule(lr){10-13} \cmidrule(lr){14-17}
& GRF & MX24 & CK & Token & GRF & MX24 & CK & Token & GRF & MX24 & CK & Token & GRF & MX24 & CK & Token \\
\midrule[1pt]
Qwen2.5-7B & 73.52 & 2.70 & 76.43 & 33.08 & 75.54 & 2.65 & 78.83 & \underline{28.33} & 69.51 & 3.23 & 76.51 & \underline{39.37} & 77.82 & 2.37 & 82.09 & 25.69 \\
Qwen2.5-14B & 74.21 & 2.65 & 77.02 & \underline{29.06} & 83.69 & 2.17 & 80.67 & 31.14 & 77.63 & 2.77 & 78.81 & 43.38 & 85.83 & 1.75 & 84.18 & 29.59 \\
Qwen2.5-14B(w/ refl) & 79.08 & 2.58 & 77.21 & 188.44 & 88.48 & 2.07 & 80.85 & 193.38 & 85.33 & 2.63 & 79.29 & 208.86 & 88.19 & 1.67 & 84.47 & 183.86 \\
Llama3.1-8B & 62.19 & 3.20 & 74.98 & 39.18 & 75.35 & 2.58 & 79.29 & 31.71 & 65.57 & 3.46 & 75.95 & 52.34 & 75.82 & 2.20 & 82.49 & 46.87 \\
\midrule[1pt]
Qwen3-8B & 79.54 & 2.53 & 77.90 & 628.63 & 88.20 & 2.00 & 81.18 & 588.59 & 82.56 & 2.59 & 79.58 & 654.06 & 89.49 & 1.61 & 84.64 & 612.31 \\
Qwen3-8B(w/ refl) & \underline{81.78} & 2.42 & \underline{78.03} & 863.12 & 89.00 & 2.00 & \textbf{81.24} & 738.92 & \underline{85.92} & 2.51 & 79.66 & 786.63 & \underline{90.78} & 1.67 & 84.72 & 726.28 \\
Marco-o1-7B & 68.43 & 2.90 & 75.80 & 489.24 & 79.58 & 2.38 & 79.42 & 548.34 & 74.67 & 2.97 & 77.35 & 584.30 & 82.72 & 1.86 & 83.50 & 535.60 \\
Marco-o1-7B(w/ refl) & 75.09 & 2.84 & 76.07 & 731.71 & 85.34 & 2.26 & 80.03 & 725.51 & 81.88 & 2.88 & 78.18 & 763.97 & 86.84 & 1.86 & 83.55 & 720.18 \\
DeepSeek-R1 & 78.00 & 3.03 & 76.50 & 541.88 & 88.37 & 2.06 & 79.39 & 593.12 & 84.03 & 2.64 & 77.93 & 605.42 & 90.23 & 1.80 & 83.17 & 783.20 \\
DeepSeek-R1(w/ refl) & 81.15 & 2.50 & 76.67 & 659.83 & \textbf{89.79} & 1.93 & 81.04 & 1029.68 & 85.61 & 2.56 & 79.28 & 1069.07 & 90.50 & 1.56 & \textbf{84.80} & 982.33 \\
QwQ-32B & 79.50 & 2.22 & 77.77 & 523.72 & 85.51 & 2.79 & 78.59 & 632.56 & 78.90 & 3.03 & 74.82 & 684.29 & 89.03 & 2.19 & 82.84 & 547.19 \\
QwQ-32B(w/ refl) & 81.65 & \underline{2.16} & 77.97 & 1063.01 & 87.17 & 2.60 & 78.33 & 1136.59 & 84.47 & 2.98 & 77.61 & 1102.00 & 89.86 & 2.11 & 82.94 & 1007.71 \\
\midrule[1pt]
DeepTrans-7B & 75.00 & 3.29 & 72.62 & 386.66 & 78.23 & 2.61 & 77.12 & 2018.96 & 76.22 & 2.99 & 77.39 & 2581.52 & 82.36 & 2.58 & 81.92 & 707.06 \\
DRT-7B & 75.69 & 2.93 & 75.48 & 478.99 & 79.70 & 2.54 & 78.87 & 479.64 & 76.39 & 3.00 & 77.40 & 520.03 & 85.15 & 2.22 & 82.85 & 463.02 \\
ExTrans-7B & 78.95 & 2.33 & 77.63 & 979.08 & 81.83 & 2.17 & 78.96 & 1547.55 & 81.70 & 2.72 & 77.51 & 1957.79 & 86.64 & 1.82 & 82.89 & 1232.43 \\
\midrule[1pt]
Qwen-7B-Adapt-SFT & 76.22 & 2.63 & 77.11 & 51.30 & 82.15 & 2.13 & 80.55 & 150.15 & 76.59 & 2.57 & 78.74 & 183.75 & 85.65 & 1.60 & 83.22 & 139.48 \\
Qwen-7B-Adapt-RL & 79.81 & 2.27 & 77.62 & 56.29 & 83.70 & \underline{1.73} & 80.77 & 192.73 & 79.40 & \underline{2.15} & \underline{79.68} & 239.31 & 86.53 & 1.37 & 84.56 & 172.99 \\
Qwen-7B-Full-RL & 78.59 & 2.32 & 77.43 & 127.47 & 82.89 & 1.74 & 80.71 & 201.31 & 79.92 & 2.16 & 79.67 & 245.99 & 86.29 & \underline{1.35} & 84.71 & 187.63 \\
\midrule
ReflectMT-SFT(Cold Start) & 76.35 & 2.65 & 76.70 & \textbf{23.25} & 86.01 & 2.21 & 80.35 & \textbf{25.94} & 80.82 & 2.73 & 79.12 & \textbf{37.57} & 87.53 & 1.70 & 84.24 & \textbf{24.09} \\
ReflectMT-Stage1(Our) & 79.73 & 2.37 & 77.45 & 29.41 & 87.89 & 1.82 & 80.78 & 28.91 & 83.39 & 2.49 & 79.41 & 41.39 & 89.32 & 1.45 & 84.39 & \underline{24.88} \\
ReflectMT-Stage2(Our) & \textbf{82.55}$^{**}$ & \textbf{2.00} & \textbf{78.05}$^{***}$ & 30.73 & \underline{89.48} & \textbf{1.70} & \underline{81.18} & 30.67 & \textbf{86.16}$^{*}$ & \textbf{2.10} & \textbf{79.86}$^{***}$ & 47.57 & \textbf{91.42} & \textbf{1.33} & \underline{84.72} & 29.10 \\
\bottomrule[1pt]
\end{tabular}
}

\label{tab:results}
\end{table*}

Our ReflectMT models employ a direct translation mode without explicit reflection during inference. Experimental results show a favorable efficiency--quality trade-off across the evaluated benchmarks. To characterize this trade-off, we compare ReflectMT with four groups of baselines:

\noindent
\textbf{Comparison with general LLMs.} As shown in Table \ref{tab:results}, our optimal model ReflectMT-Stage2 achieves a GRF score of 82.55 on our dataset, representing a 12.3\% improvement over the baseline model Qwen2.5-7B-Instruct (73.52). For MetricX-24 and CometKiwi metrics, our method achieves relative improvements of 26.0\% and 2.1\%, respectively. Notably, our model consumes an average of 30.73 tokens per sample during inference, nearly identical to the non-reflective general LLM (33.08 tokens/sample), introducing negligible additional inference overhead.

\noindent
\textbf{Comparison with strong LRM baselines.} Our model demonstrates superior performance over strong LRMs, outperforming QwQ-32B (79.50) and DeepSeek-R1 (78.00) by 3.05 and 4.55 GRF points, respectively. To ensure a rigorous comparison, we also evaluated these baselines in a ``\textit{translate-reflect-refine}'' mode. While this explicit reflection universally improves their translation quality (e.g., DeepSeek-R1 w/ refl improves to 81.15), it necessitates generating lengthy chains of thought, resulting in substantially higher token consumption (e.g., DeepSeek-R1 consumes 541.88 tokens/sample, 17$\times$ that of our model). In contrast, our model, built upon the compact Qwen2.5-7B backbone, internalizes this reflection capability to generate high-quality translations directly, offering significant advantages in both parameter scale and computational efficiency.

\noindent
\textbf{Comparison with MT-specialized models.} On our in-house test set, ReflectMT achieves a GRF score of 82.55, compared with 78.95 for ExTrans-7B, while generating substantially fewer output tokens. Results on other benchmarks and metrics demonstrate its appealing efficiency–quality trade-off.

\noindent
\textbf{Comparison with pre-thinking models.} To verify the advantages of post-thinking reflection over pre-thinking, we trained a pre-thinking baseline under identical training configurations. Compared to the adaptive pre-thinking model (79.81), our method improves by 2.74 GRF points while reducing token consumption by 45\%. On this evaluation setting, these results suggest that the post-thinking design can improve the efficiency--quality trade-off relative to the adaptive pre-thinking baseline.

We additionally conducted a blinded human evaluation on 150 examples
from our in-house test set; the evaluation protocol and results
are provided in Appendix~\ref{app:Human Evaluation}.

\subsection{Human Evaluation}
\label{subsec:human-evaluation}

To complement automatic evaluation, we randomly sampled 150 source sentences from our in-house test set and compared ReflectMT with DeepSeek-R1 with reflection. Two bilingual NLP researchers independently scored the translations on a 0--100 scale, considering semantic faithfulness, fluency, and adequacy relative to the source. System identities were hidden and the presentation order was randomized.

\begin{table}[htbp]
    \centering
    \small
    \caption{Blinded human evaluation on 150 in-house test sentences. Scores are averaged over two annotators (0--100; higher is better).}
    \label{tab:human-eval-main}
    \begin{tabular}{lc}
        \toprule
        System & Human score \\
        \midrule
        DeepSeek-R1 (w/ refl) & 89.99 \\
        ReflectMT & \textbf{91.34} \\
        \bottomrule
    \end{tabular}
\end{table}

As shown in Table~\ref{tab:human-eval-main}, ReflectMT obtains an average score of 91.34, exceeding DeepSeek-R1 with reflection by 1.35 points. Both annotators assign higher average scores to ReflectMT. These results complement the automatic metrics, suggesting that our efficient first-pass translations retain high quality under human assessment on this sampled subset. Per-annotator scores and agreement analysis are provided in Appendix~\ref{app:Human Evaluation}.

\subsection{Reflection Internalization Analysis}
\label{subsec:3.4}

\begin{figure*}[h]
    \centering
    \subfloat[Number of Modifications]{\includegraphics[width=0.48\linewidth]{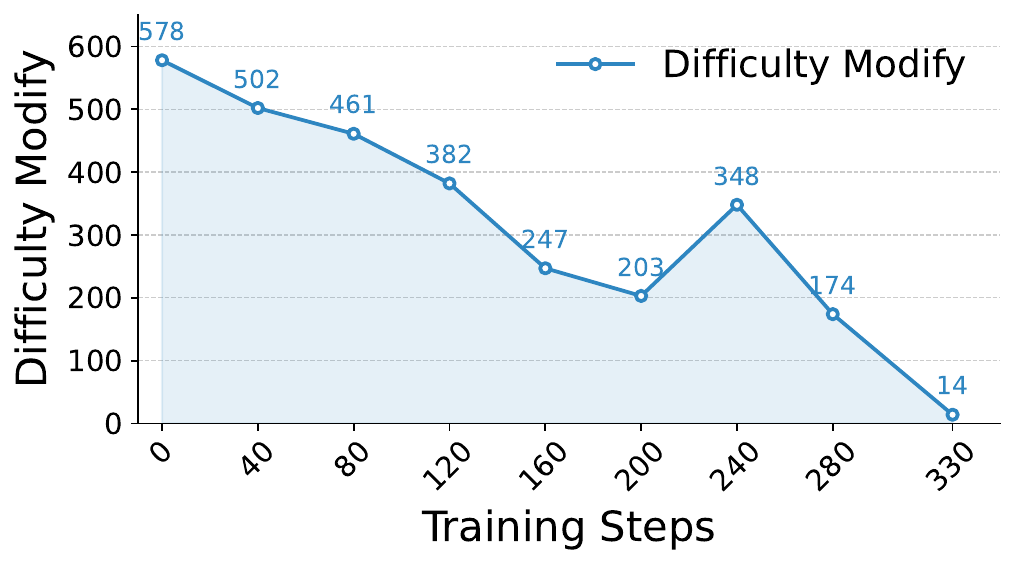}}
    \subfloat[Accuracy]{\includegraphics[width=0.48\linewidth]{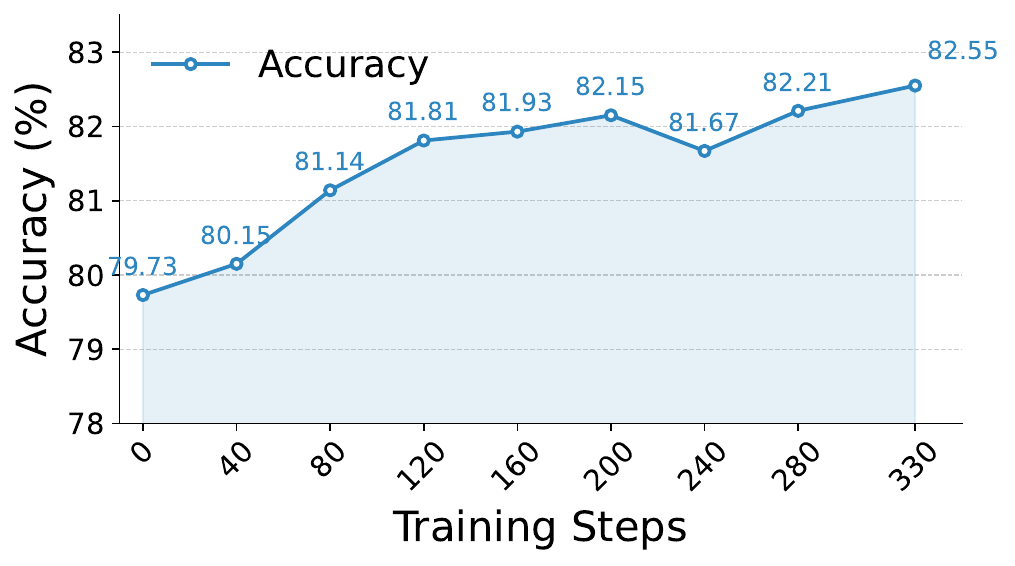}}     \\
    \subfloat[Divergence]{\includegraphics[width=0.48\linewidth]{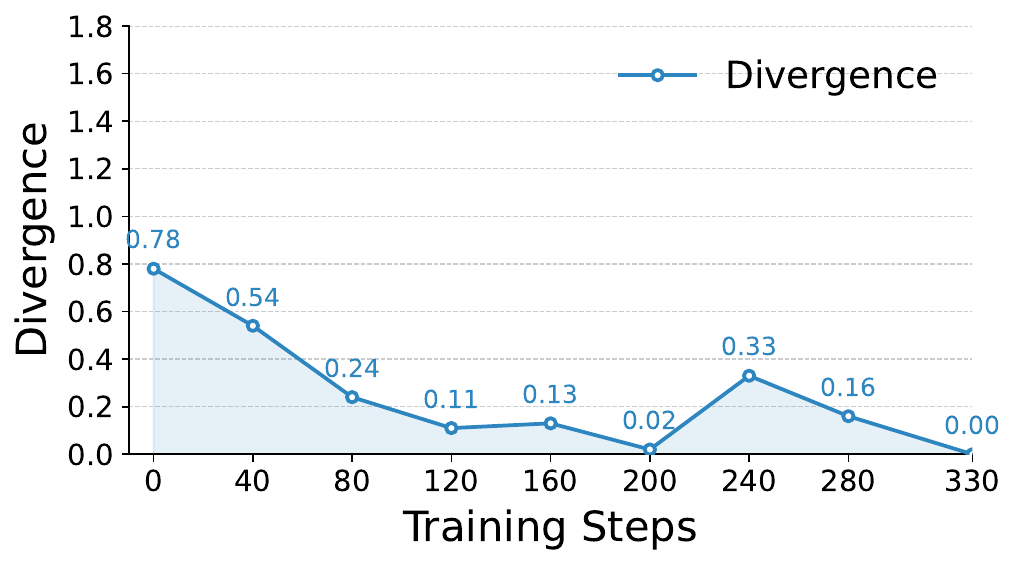}}
    \subfloat[Modifications by Difficulty]{ \includegraphics[width=0.48\linewidth]{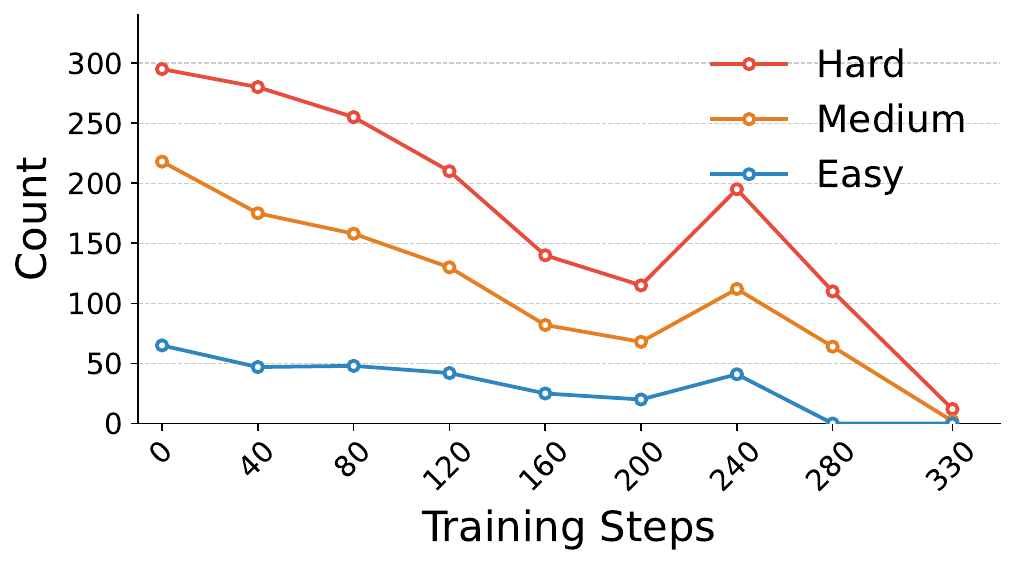}}
    \caption{Training Dynamics: As the training steps increase (a) indicates the number of refinements; (b) represents the GRF score; (c) illustrates the score difference between initial and final translations; (d) shows the number of refinements for tasks of varying difficulty.}
\label{fig:intermediate_analyses}
\end{figure*}

\textbf{Training Dynamics of Reflection Internalization.}
We evaluate the model every 40 steps during Stage 2 on a 2000-sample test set, temporarily disabling early-stopping to observe both initial and final translations. As shown in Figure \ref{fig:intermediate_analyses}(b), GRF scores exhibit a fluctuating upward trend, reflecting the model's adjustment of internal representations to balance preliminary translation quality and reflective improvement ability during the gradual internalization of its reflective capability.

\begin{table*}[h]
\centering

\caption{Comparative performance of initial translations and final refinements across different training stages. The shrinking performance gap demonstrates the successful internalization of reflection capabilities.}
\resizebox{\textwidth}{!}
{
\begin{tabular}{lcccccccccccccccc}
\toprule[1pt]
\multirow{2}{*}{Model} & \multicolumn{4}{c}{Our} & \multicolumn{4}{c}{WMT23} & \multicolumn{4}{c}{WMT24} & \multicolumn{4}{c}{FLORES-200} \\
\cmidrule(lr){2-5} \cmidrule(lr){6-9} \cmidrule(lr){10-13} \cmidrule(lr){14-17}
& GRF & MX24 & CK & Token & GRF & MX24 & CK & Token & GRF & MX24 & CK & Token & GRF & MX24 & CK & Token \\
\midrule[1pt]
ReflectMT-SFT-initial & 76.35 & 2.65 & 76.70 & \textbf{23.25} & 86.01 & 2.21 & 80.35 & \textbf{25.94} & 80.82 & 2.73 & 79.12 & 37.57 & 87.53 & 1.70 & 84.24 & \textbf{24.09} \\
ReflectMT-SFT-final & 77.26 & 2.63 & 77.32 & 297.51 & 86.05 & 2.16 & 80.87 & 273.78 & 83.15 & 2.69 & 79.42 & 323.75 & 87.83 & 1.69 & 84.30 & 248.34 \\
ReflectMT-Stage1-initial & 79.73 & 2.37 & 77.45 & \underline{29.41} & 87.89 & 1.82 & 80.78 & \underline{28.91} & 83.39 & 2.49 & 79.41 & 41.39 & 89.32 & 1.45 & 84.39 & \underline{24.88} \\
ReflectMT-Stage1-final & 80.51 & 2.23 & 77.81 & 261.25 & 88.46 & 1.87 & 80.99 & 239.15 & 84.55 & 2.24 & 79.51 & 298.27 & 90.06 & 1.39 & 84.50 & 322.28 \\
ReflectMT-Stage2-initial & \textbf{82.55} & \textbf{2.00} & \textbf{78.05} & 30.73 & \underline{89.48} & \textbf{1.70} & \textbf{81.18} & 30.67 & \underline{86.16} & \textbf{2.10} & \underline{79.86} & 47.57 & \underline{91.42} & \textbf{1.33} & \textbf{84.72} & 29.10 \\
ReflectMT-Stage2-final & \underline{82.55} & \underline{2.00} & \underline{78.05} & 322.29 & \textbf{89.49} & \underline{1.70} & \underline{81.18} & 288.35 & \textbf{86.19} & \underline{2.10} & \textbf{79.87} & 350.77 & \textbf{91.47} & \underline{1.33} & \underline{84.72} & 244.50 \\
\bottomrule[1pt]
\end{tabular}
}

\label{tab:i-f}
\end{table*}

Furthermore, we monitor the frequency of explicit modifications. Crucially, the model learns to self-assess, triggering refinement only when it judges the initial translation as sub-optimal. As shown in Figure \ref{fig:intermediate_analyses}(a), modifications drop monotonically from 578 (28.9\%) at Step 0 to a mere 14 (0.7\%) by Step 330. This trend indicates that as the reflective capability is internalized, the model generates higher-quality translations in the first pass. Figure \ref{fig:intermediate_analyses}(c) shows that the score difference between first-pass and final translations exhibits a fluctuating downward trend. As the reflective capability is gradually internalized, the model ultimately generates optimal translations in the first pass without subsequent modifications. Table \ref{tab:i-f} displays first-pass and final translation results at different training stages. This demonstrates the effectiveness of our reflective internalization training framework: the model successfully incorporates the knowledge learned during reflection into its initial translation capability, achieving quality close to the final translation during the first generation.

\noindent
\textbf{Performance Across Difficulty Levels.}
To evaluate model performance on translation tasks of varying difficulty, we divided samples into three levels based on GRF scores: easy (>90), medium (70-90), and difficult (<70), with proportions approximately 11:6:3. We recorded the number of samples requiring reflective modifications during training at each difficulty level.

As shown in Figure \ref{fig:intermediate_analyses}(d), at Step 0, the model modified 65 easy, 218 medium, and 295 difficult samples, indicating significant room for improvement in the initial translation stage despite acquired reflective capability after first-phase training, especially for difficult samples.

After second-phase reflective internalization training, model performance at Step 330 showed qualitative improvement. For easy samples, the model required no reflective modifications, indicating first-pass translations meet high-quality standards. For difficult samples, modifications drastically decreased from 295 to 12, a 96\% reduction. This demonstrates the model's sensitivity to translation difficulty: for easy and medium samples, the model fully internalized reflective capability, generating high-quality translations during initial translation; for genuinely difficult samples, the model retained reflective improvement capability to ensure translation quality.

\subsection{Ablation Study}
\label{subsec:3.5}

\begin{table}[h]
\centering
\caption{Ablation study on the reflection mechanism evaluated by GRF scores. ``\textit{Initial}'' and ``\textit{Final}'' refer to the translations generated before and after explicit reflection. The ``\textit{RL}'' stage is initialized from the checkpoint of the SFT model. }
\resizebox{0.48\textwidth}{!}
{
\begin{tabular}{lcc}
\toprule[1pt]
\multicolumn{1}{c}{Model Configuration} & Initial & Final \\
\midrule[1pt]
Qwen-7B-Instruct (Baseline) & 73.52 & -- \\
ReflectMT-SFT (w/o init \& refl) & -- & 75.86 \\
ReflectMT-SFT (w/o refl) & \underline{76.85} & 76.49 \\
ReflectMT-SFT (full) & 76.35 & 77.26 \\
ReflectMT-RL (w/o init \& refl) & -- & \underline{78.62} \\
ReflectMT-RL (full) & \textbf{79.73} & \textbf{80.51} \\
\bottomrule[1pt]
\end{tabular}
}

\label{tab:ablation}
\end{table}

To validate the effectiveness of the reflection mechanism, we compare three training configurations, where \textit{init} and \textit{refl} denote initial translation and reflection:

\textit{1)ReflectMT (w/o init \& refl).} The model is trained to map the source sentence directly to the final translation, omitting both the initial translation and the reflection steps.

\textit{2)ReflectMT (w/o refl).} The model is trained to generate an initial translation followed directly by a refined translation, bypassing the explicit reflection analysis. This variant learns a ``\textit{translation-refinement}'' process.

\textit{3)ReflectMT (full).} Our complete configuration, which trains the model on the full ``\textit{translation-reflection-refinement}'' trajectory.

Table \ref{tab:ablation} demonstrates that the reflection mechanism significantly enhances translation quality. During the SFT stage, the \textit{w/o refl} variant yields a final GRF score (76.49) lower than its initial translation (76.85), indicating that blind refinement without explicit guidance can degrade performance. Conversely, the \textit{full} model improves from 76.35 to 77.26, proving reflection provides accurate modification directions. Having established the necessity of reflection during SFT, we evaluate the ultimate performance ceiling after RL training. Comparing direct translation (\textit{w/o init \& refl}, 78.62) to the full ReflectMT (80.51), the 1.89-point GRF gap confirms the substantial benefit of the complete ``\textit{translate-reflect-refine}'' paradigm.

\section{Related Work}
\vspace{0.5ex}
\noindent \textbf{Application of Deep Reasoning Models in Machine Translation.} With the emergence of Large Reasoning Models (LRMs) such as OpenAI o1~\cite{openai_o1_2024} and DeepSeek-R1 ~\cite{guo2025deepseek}, deep reasoning capabilities have pioneered extensive research in long Chain-of-Thought (CoT) reasoning. Recently, researchers have begun exploring their potential in Machine Translation (MT). \citet{zhao2024marco} and \citet{liu2025new} demonstrated the prospects of long CoT reasoning in translation through heuristic examples. \citet{wang2024drt} constructed the MetaphorTrans dataset and trained the DRT model to handle rhetorical devices in literary translation. DeepTrans-7B~\cite{wang2025deep} leveraged DeepSeek-V3 for reference-free evaluation during Reinforcement Learning (RL) training, exploiting the powerful capability of large language models as judges. Building on this, \citet{wang2025extrans} proposed the ExTrans model with exemplar-enhanced RL reward modeling, integrating dual capabilities of LLMs as both exemplar generators and judges, achieving breakthroughs in deep reasoning MT and extending to low-resource multilingual translation.

\vspace{0.5ex}
\noindent \textbf{Reflection and Self-Improvement.}
Reflection and self-improvement have been widely explored to enhance LLM outputs.
\citet{shinn2023reflexion} proposed Reflexion, which converts task feedback into verbal reflections and stores them in episodic memory to guide subsequent attempts without updating model parameters.
\citet{ji2023mitigatinghallucinationlargelanguage} explored iterative self-reflection to mitigate hallucinations by checking the factuality of generated knowledge and the consistency of answers.
\citet{weng2023largelanguagemodelsbetter} introduced backward verification, treating candidate conclusions as conditions and using verification scores to select reliable answers.

Recent work has further improved the applicability and reliability of reflection.
TasTe~\cite{wang2024tasteteachinglargelanguage} applies reflection to translation through multi-task supervised instruction tuning for translation, quality prediction, and draft refinement.
Its two-stage inference procedure first generates a draft and predicts its quality as a label or score, then refines the translation conditioned on the source text, draft, and quality prediction.
IoRT~\cite{liu2025instructofreflectionenhancinglargelanguage} addresses redundancy, drift, and persistent errors in iterative reflection through dynamic meta-instructions.
An instructor guided by meta-thoughts and answer-consistency information issues refresh, stop, or select instructions to regulate subsequent iterations.
PR-CoT~\cite{costa2026enhancingselfcorrectionlargelanguage} uses prompting to organize structured reflection across logical consistency, information completeness, potential biases, and alternative solutions, then synthesizes these critiques to refine the initial reasoning and answer without additional training.

Unlike approaches that retain explicit reflection or verification at inference time, ReflectMT aims to internalize reflection benefits through two-stage reinforcement learning.
In particular, its Stage-2 reward optimization targets first-pass translation rather than relying on assessment-conditioned refinement.
This shifts the focus from executing a reflective inference procedure to improving a direct translation policy, without explicit reasoning or iterative revision at deployment.

\vspace{0.5ex}
\noindent \textbf{Knowledge Distillation and Mutual Learning.}
Knowledge distillation has been explored in MT through confidence-based transfer of bidirectional global-context information~\cite{zhou-etal-2022-confidence}, joint teacher--student learning~\cite{miao2023exploring}, and future-context transfer~\cite{zhang2019future}. 
Related work in code generation exploits complementary AST traversal orders through mutual distillation~\cite{xie2021improving}.
These studies leverage additional knowledge during training, while ReflectMT focuses on internalizing reflection through RL for direct translation.
SPEC-Distill~\cite{liu-etal-2026-spec} further suggests a potential direction for reducing our RL rollout costs through speculative prefix reuse, which we leave for future work.

\section{Conclusion}
This paper introduces a novel ``\textit{translate-reflect-refine}'' paradigm for machine translation. We design a multi-agent collaborative system to construct reflective translation datasets and propose a two-stage reinforcement learning framework that internalizes reflection capabilities into the model's direct responses.
Across our primary English-to-Chinese setting and preliminary evaluations on three additional directions, ReflectMT consistently reduces inference-time output tokens relative to explicit-reflection LRMs while achieving competitive translation quality. Critically, our model generates high-quality translations without explicit reflection steps during inference, avoiding the computational overhead of long COT reasoning while maintaining competitive translation quality. Ablation studies further confirm the effectiveness of the reflection mechanism in our framework.

\section{Limitations} 
\label{limit}

While we have demonstrated the effectiveness of our reflection framework across multiple language pairs, several limitations are worth noting. Evaluation on low-resource languages remains limited, and the effectiveness of our reflection mechanism under scarce training data and limited linguistic resources requires further empirical investigation. Moreover, while our model significantly reduces computational overhead during inference, the training pipeline—including multi-agent data construction and two-stage RL—introduces certain upfront computational costs compared to standard fine-tuning.

\section{Acknowledgements}

The project was supported by National Natural Science Foundation of China (No. 62676342), the Open Competition for Innovative Projects of Xiamen (No. 3502Z20251012), Natural Science Foundation of Fujian Province of China (No. 2024J011001), and the Public Technology Service Platform Project of Xiamen (No.3502Z20231043). We also thank the reviewers for their insightful comments.

\newpage
\medskip
{
    \small
    \bibliographystyle{plainnat}
    \bibliography{neurips}
}


\appendix

\section{Our Data}
\label{app:Data}

Our dataset is constructed by randomly sampling from WMT19~\cite{ma2019results}, SlangDIT~\cite{liang2025slangditbenchmarkingllmsinterpretative}, and MetaphorTrans~\cite{wang2024drt}. We employ DeepSeek-V3 to generate the corresponding translations and reflections. Subsequently, we categorize the difficulty of these samples based on their final GRF scores and filter the data accordingly. The resulting dataset maintains an approximate 1:1 ratio of simple to difficult samples. The detailed dataset composition and the numbers of training, validation, and test examples from each source are reported in Table~\ref{tab:dataset} and Table~\ref{tab:source-splits}, respectively, while Table~\ref{tab:round-distribution} presents the distribution of reflection rounds in the training set. By incorporating these difficult samples, we aim to fully elicit the model's reasoning capabilities, thereby encouraging it to learn in-depth reflection and idiomatic translation. 

During the Cold Start Supervised Fine-Tuning (SFT) stage, we train the model using the complete samples---comprising the initial translation, reflection analysis, and final translation---to help it internalize the entire reflective translation pipeline. Specifically, rather than retaining only the last refinement transition, we place the initial translation $y_0$ in \texttt{<answer>}, use an LLM to consolidate the feedback and scores from subsequent rounds into a single comprehensive \texttt{<reflection>}, and place the final translation $y_K$ in \texttt{<final\_answer>}. Each SFT instance therefore captures the trajectory from the initial draft to the final refinement, even when the data-construction loop involves multiple reflection rounds. Conversely, in the Reinforcement Learning (RL) stage, we provide only the source English sentences as input, without any reference translations or annotations. For data generation, we locally deploy DeepSeek-V3 (671B) with the temperature set to 0.1 and top-$p$ to 0.95. To ensure data quality, we use rule-based rejection sampling that combines regular-expression format checks with predefined rejection criteria applied to LLM Reflector scores, rather than relying on an LLM to make filtering decisions directly. Furthermore, the hyperparameters in Section~\ref{subsec:2.2}(3) are set as follows: $\theta=90$, $K_{max}=5$, $r_{low}=40$.

The data format for cold start SFT is as follows:

\begin{tcolorbox}[
    colback=gray!10!white,     
    colframe=black,            
    boxrule=1pt,               
    boxsep=4pt,                
    top=4pt, bottom=4pt,       
    left=8pt, right=8pt        
]
\begin{CJK}{UTF8}{gbsn}
<answer>2018年非洲面临的三大人道主义挑战</answer> <reflection>1. 翻译质量分析：- 准确性：翻译准确传达了原文的核心信息，"Three Humanitarian Challenges"被准确地译为"三大人道主义挑战"，"for Africa in 2018"也被正确地处理为"2018年非洲面临的"。   - 流畅性：译文符合中文表达习惯，语句通顺自然。   - 语境适配：译文很好地适应了新闻报道或学术文章的语境，保持了正式和客观的语气。   - 过度翻译：未发现过度翻译现象，译文简洁明了。2. 存在的问题：   - 无显著问题。译文在各方面都达到了较高的标准。3. 改进建议：  - 虽然译文已经很完美，但可以考虑在特定语境下增加一些修饰词以增强感染力，例如："2018年非洲亟待解决的三大紧迫人道主义挑战"。这种调整可以根据具体使用场景灵活选择。"</reflection> <need\_revision>否</need\_revision> <final\_answer>2018年非洲面临的三大人道主义挑战</final\_answer>
\end{CJK}
\end{tcolorbox}

The initial format of our reflective translation dataset is as follows:
\begin{tcolorbox}[
    colback=gray!10!white,     
    colframe=black,            
    boxrule=1pt,               
    boxsep=4pt,                
    top=4pt, bottom=4pt,       
    left=8pt, right=8pt        
]

\begin{CJK}{UTF8}{gbsn}\{"src": "The big losers from this economic disaster are workers in the advanced countries that bought into the laissez-faire flexibility of American-style capitalism.", 

"ref": "这场经济灾难的大输家是那些认同美式资本主义放任自流政策的发达国家的工人。", 

"final\_translation": "在这场经济灾难中遭受重大损失的是那些接受了美国式资本主义自由放任灵活性的发达国家的工人。",

"total\_rounds": 1, 

"final\_score": 90, 

"best\_round": 1, 

"history": [\{"round": 1, 

"translation": "在这场经济灾难中遭受重大损失的是那些信奉美国式资本主义自由放任灵活性的发达国家的工人。", 

"reflection": "1. 翻译质量分析：   - 准确性：翻译基本准确传达了原文的意思，特别是“economic disaster”（经济灾难）、“big losers”（遭受重大损失）和“American-style capitalism”（美国式资本主义）等关键术语的处理得当。   - 流畅性：译文整体较为流畅，符合中文表达习惯。句子结构清晰，逻辑连贯。   - 语境适配：译文很好地适应了原文的经济学背景，尤其是对“laissez-faire flexibility”（自由放任灵活性）的解释和处理，符合经济学领域的常用表达。   - 过度翻译：没有明显的过度翻译现象，所有信息点都得到了恰当的传达。2. 存在的问题：   - “bought into”这一短语的翻译略显生硬，直译为“信奉”虽然可以理解，但可能不够精准。“bought into”在这里更强调对这些理念的接受或认同，而非单纯的“信奉”。3. 改进建议：   - 可以将“bought into”翻译为“接受了”或“认同了”，以更准确地传达原文的含义。例如：“在这场经济灾难中遭受重大损失的是那些接受了美国式资本主义自由放任灵活性的发达国家的工人。", 

"score": 90\}]\}\}
\end{CJK}
\end{tcolorbox}

\begin{table}[htbp]
\centering
\small
\caption{Per-source dataset statistics.}
\begin{tabular}{lrrr}
\toprule
Source & Train & Validation & Test \\
\midrule
WMT19         & 5,482  & 400   & 866   \\
SlangDIT      & 2,856  & 259   & 497   \\
MetaphorTrans & 3,051  & 341   & 637   \\
\midrule
Total         & 11,389 & 1,000 & 2,000 \\
\bottomrule
\end{tabular}
\label{tab:source-splits}
\end{table}

\begin{table}[htbp]
\centering
\small
\caption{Distribution of \texttt{total\_rounds} in the training set.}
\begin{tabular}{crr}
\toprule
total\_rounds & Instances & Percentage \\
\midrule
1 & 3,443 & 30.23\% \\
2 & 5,425 & 47.63\% \\
3 &   778 &  6.83\% \\
4 &   362 &  3.18\% \\
5 & 1,381 & 12.13\% \\
\bottomrule
\end{tabular}
\label{tab:round-distribution}
\end{table}

\noindent
\textbf{Initial-Translator Robustness.} Both roles use DeepSeek-V3 in our default data-construction pipeline. As a model-separation check, we replace only the initial Translator with Qwen-72B while keeping DeepSeek-V3 as the Reflector. Table~\ref{tab:translator-separation} reports the GRF scores of the downstream cold-start SFT models under both configurations. This comparison provides preliminary evidence of robustness to the initial-translator choice, rather than a claim that using the same model is universally optimal.

\begin{table}[htbp]
    \centering
    \caption{Effect of the initial-translator choice on downstream
    cold-start SFT performance. Translator and Reflector denote
    the models used for initial translation and reflection during
    data construction, respectively.}
    \label{tab:translator-separation}
    \begin{tabular}{llc}
        \toprule
        Translator & Reflector & SFT GRF \\
        \midrule
        DeepSeek-V3 & DeepSeek-V3 & 76.35 \\
        Qwen-72B    & DeepSeek-V3 & 75.83 \\
        \bottomrule
    \end{tabular}
\end{table}


\section{Human Evaluation Details}
\label{app:Human Evaluation}
Table~\ref{tab:human-eval} provides the per-annotator mean scores underlying the aggregate results in Section~\ref{subsec:human-evaluation}. Inter-annotator agreement, measured using the two-way random-effects, absolute-agreement intraclass correlation coefficient for average ratings (ICC(2,k))~\citep{article}, was 0.58, indicating moderate agreement.

\begin{table}[t]
    \centering
    \caption{Per-annotator mean scores for the human evaluation in Section~\ref{subsec:human-evaluation}.}
    \label{tab:human-eval}
    \begin{tabular}{lcc}
        \toprule
        Annotator & DeepSeek-R1 (w/ refl) & ReflectMT \\
        \midrule
        1 & 90.82 & 91.40 \\
        2 & 89.15 & 91.28 \\
        \bottomrule
    \end{tabular}
\end{table}

\section{Robustness Analysis}
\label{app:Robustness Analysis}

\textbf{Effect of Reward Judge Choice.} To examine reward-judge dependence, we repeated RL training with Qwen-72B replacing DeepSeek-V3 as the reward judge. Table~\ref{tab:reward-judge} compares the resulting Stage-2 model with the model trained using the original reward judge and the DeepSeek-R1 baseline with reflection. These results suggest that the efficiency--quality trade-off is not solely tied to the DeepSeek-V3 reward judge, while showing some quality variation across judge choices.

\begin{table*}[t]
    \centering
    \caption{Reward-judge robustness. The first two rows are reproduced
    from Table~\ref{tab:results}; the last row replaces DeepSeek-V3 with Qwen-72B as
    the reward judge during RL training.}
    \label{tab:reward-judge}
    \resizebox{\textwidth}{!}{%
    \begin{tabular}{l*{16}{r}}
        \toprule
        & \multicolumn{4}{c}{Our}
        & \multicolumn{4}{c}{WMT23}
        & \multicolumn{4}{c}{WMT24}
        & \multicolumn{4}{c}{FLORES-200} \\
        \cmidrule(lr){2-5}
        \cmidrule(lr){6-9}
        \cmidrule(lr){10-13}
        \cmidrule(lr){14-17}
        Model
        & GRF & MX24 & CK & Token
        & GRF & MX24 & CK & Token
        & GRF & MX24 & CK & Token
        & GRF & MX24 & CK & Token \\
        \midrule
        DeepSeek-R1 (w/refl)
        & 81.15 & 2.50 & 76.67 & 659.83
        & 89.79 & 1.93 & 81.04 & 1029.68
        & 85.61 & 2.56 & 79.28 & 1069.07
        & 90.50 & 1.56 & 84.80 & 982.33 \\
        ReflectMT-Stage2 (Our)
        & 82.55 & 2.00 & 78.05 & 30.73
        & 89.48 & 1.70 & 81.18 & 30.67
        & 86.16 & 2.10 & 79.86 & 47.57
        & 91.42 & 1.33 & 84.72 & 29.10 \\
        ReflectMT-Stage2 (Qwen-72B reward)
        & 81.91 & 2.26 & 78.92 & 28.94
        & 88.72 & 1.98 & 81.11 & 29.63
        & 85.63 & 2.41 & 79.83 & 49.16
        & 90.97 & 1.53 & 84.51 & 29.95 \\
        \bottomrule
    \end{tabular}%
    }
\end{table*}

\textbf{Multi-Seed RL Results.} A single RL run may not provide sufficient evidence of a reliable performance advantage. We therefore conducted two additional RL runs with different random seeds during the RL stage. Together with the original run, this yields three independent runs with WMT24 GRF scores of 86.16, 86.05, and 85.97, corresponding to a mean of 86.06 and a sample standard deviation of 0.10. For comparison, Qwen3-8B with reflection and DeepSeek-R1 with reflection achieve WMT24 GRF scores of 85.92 and 85.61, respectively. These results suggest that our method consistently achieves competitive translation quality compared with existing large reasoning models (LRMs), rather than relying on a favorable outcome from a single run.

\textbf{Statistical Significance Tests.} We selected DeepSeek-R1 with reflection as a representative large reasoning model (LRM) baseline for statistical comparison. To assess whether the observed performance differences are statistically significant, we conducted paired bootstrap tests~\cite{tibshirani1993introduction} on WMT24 ($N=998$) and our evaluation dataset ($N=2000$). As shown in Table~\ref{tab:bootstrap}, the GRF differences are statistically significant on both datasets, with $p=0.015$ on WMT24 and $p=0.0053$ on our evaluation dataset. The CometKiwi (CK) differences are also significant on both datasets ($p<0.001$). These results provide statistical support for the observed performance differences relative to DeepSeek-R1 with reflection.

\begin{table}[t]
\centering
\small
\caption{Paired bootstrap tests against DeepSeek-R1 with reflection. Significance levels: * p < 0.05,
** p < 0.01, and *** p < 0.001.}
\begin{tabular}{lllrrc}
\toprule
Dataset & Baseline & Metric & $N$ & $p_{\mathrm{boot}}$ & Sig. \\
\midrule
WMT24 & DeepSeek-R1 w/ refl & GRF & 998  & 0.015    & *   \\
WMT24 & DeepSeek-R1 w/ refl & CK  & 998  & $<0.001$ & *** \\
\midrule
Our   & DeepSeek-R1 w/ refl & GRF & 2000 & 0.0053   & **  \\
Our   & DeepSeek-R1 w/ refl & CK  & 2000 & $<0.001$ & *** \\
\bottomrule
\end{tabular}

\label{tab:bootstrap}
\end{table}

\section{Evaluation Prompt}
\label{app:Evaluation Prompt}

The evaluation prompt of GRF borrows from \citet{kocmi-federmann-2023-large}, and is also employed in \citet{wang2024drt, wang2025deep, wang2025extrans}:
\begin{tcolorbox}[
    colback=gray!10!white,     
    colframe=black,            
    boxrule=1pt,               
    boxsep=4pt,                
    top=4pt, bottom=4pt,       
    left=8pt, right=8pt        
]
Score the following translation from \texttt{[src lang]} to \texttt{[trg lang]} on a continuous scale from 0 to 100, where score of zero means "no meaning preserved" and score of one hundred means "perfect preservation of meaning, with faithfulness, expressiveness, and elegance".\\

\texttt{[src lang]} source: \{src\}\\
\texttt{[trg lang]} translation: \{hyp\}\\

Score: 

\end{tcolorbox}
The prompt evaluates a translation from a general perspective, and achieves a high correlation with humans~\cite{kocmi-federmann-2023-large}.

\section{Data Construction Prompts}
\label{app:Data Construction Prompts}
The Translator uses a minimal translation instruction without a specialized persona; translation-specific feedback is provided through the Reflector and refinement instructions. The exact prompts for all three steps, including the required output tags, are provided below.

\textbf{Translator Prompt.} 

\begin{tcolorbox}[
    colback=gray!10!white,     
    colframe=black,            
    boxrule=1pt,               
    boxsep=4pt,                
    top=4pt, bottom=4pt,       
    left=8pt, right=8pt        
]
Please translate the following English text into Chinese. Only provide the translation result:

Source text: {src\_text}

Please return your answer in the following format:

<answer>translation</answer>

\end{tcolorbox}

\textbf{Reflector Prompt.} 

\begin{tcolorbox}[
    colback=gray!10!white,     
    colframe=black,            
    boxrule=1pt,               
    boxsep=4pt,                
    top=4pt, bottom=4pt,       
    left=8pt, right=8pt        
]

Please reflect on and evaluate the following English-to-Chinese translation:

Source text (English): {src\_text}

Current translation (Chinese): {translation}

Previous reflection: {previous\_reflection}

Please evaluate the translation from the perspectives of accuracy, fluency, and contextual adequacy, and provide suggestions for improvement along with a score.

Please strictly return your answer in the following format:
<reflect>

Detailed reflection, including:

1. Translation quality analysis (accuracy, fluency, contextual adequacy, whether there is over-translation)

2. Existing problems (if any)

3. Suggestions for improvement (if any)

</reflect>

<score>score (an integer between 0 and 100)</score>

\end{tcolorbox}

\textbf{Refinement Prompt.} 

\begin{tcolorbox}[
    colback=gray!10!white,     
    colframe=black,            
    boxrule=1pt,               
    boxsep=4pt,                
    top=4pt, bottom=4pt,       
    left=8pt, right=8pt        
]

Please improve the following English-to-Chinese translation based on the reflection:

Source text (English): {src\_text}

Current translation (Chinese): {previous\_translation}

Reflection: {reflection}

Please provide the improved translation in the following format:

<answer>improved translation</answer>

\end{tcolorbox}

\section{Reflection Quality Evaluation Prompt}

We use DeepSeek-V3 to evaluate reflection quality based on accurate issue identification, in-depth analysis, justified alternatives, and coherent reasoning. The exact evaluation prompt is provided below, and the resulting score on a 0--100 scale is normalized to $[0,1]$ to serve as the reflection reward.

\begin{tcolorbox}[
    colback=gray!10!white,     
    colframe=black,            
    boxrule=1pt,               
    boxsep=4pt,                
    top=4pt, bottom=4pt,       
    left=8pt, right=8pt        
]
You are an expert in evaluating translation reflection processes.
A translation task requires translating the following English text into Chinese:

<source>{source}</source>

Someone provided an initial translation:

<initial\_translation>{initial\_translation}</initial\_translation>
Then they reflected on their translation:

<reflection>{reflection}</reflection>

Please evaluate the quality of the reflection process on a scale of 0 to 100, where:

- **0 points**: Empty reflection or less than 20 characters; no meaningful content.

- **1-10 points**: Extremely superficial reflection; only generic statements like "translation is correct" with no actual analysis.

- **11-30 points**: Basic reflection; mentions some issues but lacks depth; no discussion of nuances, context, or alternative strategies.

- **31-50 points**: Moderate reflection; identifies some translation issues and provides basic analysis; mentions alternative approaches but without detailed justification.

- **51-70 points**: Good reflection; demonstrates deep analysis of translation issues, context, and cultural nuances; discusses alternative strategies with reasonable justification.

- **71-100 points**: Excellent reflection; provides comprehensive analysis of translation problems, explores multiple alternatives with detailed justification; shows deep understanding of both languages and cultural contexts.

**Evaluation Criteria**:

1. **Problem Identification** (30\%): Does it accurately identify issues in the initial translation?

2. **Analysis Depth** (30\%): Does it deeply analyze why the initial translation has problems?

3. **Alternative Solutions** (25\%): Does it propose better translation alternatives?

4. **Logical Coherence** (15\%): Is the reflection process clear and well-organized?

Only output a score.

score: Your score
\end{tcolorbox}

\begin{table}[t]
\centering
\caption{Dataset statistics for translation tasks.}

\begin{tabular}{lcc}
\toprule[1pt]
Split & EN$\rightarrow$ZH & ZH$\rightarrow$EN \\
\midrule[1pt]
Train & 11,389 & 8,000 \\
Test & 2,000 & 2,000 \\
Val & 1,000 & 1,000 \\
\bottomrule[1pt]
\end{tabular}
\label{tab:dataset}
\end{table}

\begin{figure}[t]
    \centering
    \subfloat[Stage 1]
    {\includegraphics[width=0.48\linewidth]{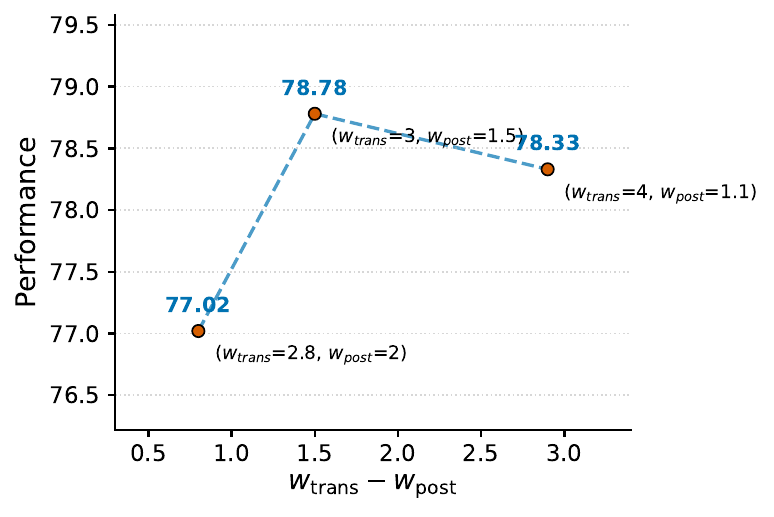}}
    \subfloat[Stage 2]{\includegraphics[width=0.48\linewidth]{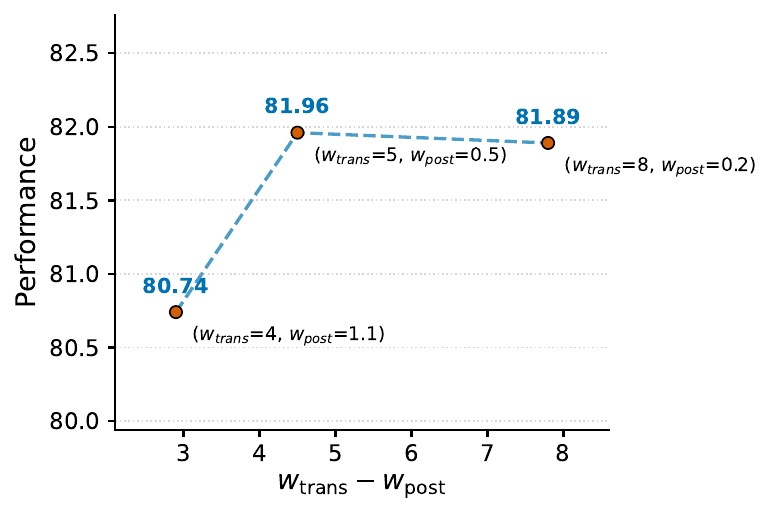}}
    \caption{Effect of the reward-weight gap $w_{\mathrm{trans}}-w_{\mathrm{post}}$ on the two training stages. Scores are reported on the validation set.}
    \label{fig:hp_analysis}
\end{figure}

\section{Implementation Details}
\label{app:Details}

\textbf{Cold Start SFT.} We employed the LLaMA-Factory framework for Supervised Fine-Tuning (SFT). The model was trained for 3 epochs with a learning rate of $1\times 10^{-4}$ and a batch size of 1 with 8 gradient accumulation steps. The entire SFT process required approximately 3 GPU hours (equivalent NVIDIA A100 40G GPU hours).

\textbf{RL Training.} We implemented the GRPO algorithm using the \texttt{verl}\footnote{\url{https://github.com/volcengine/verl}} library. We locally deployed DeepSeek-V3 to serve as the reward model for optimizing the policy model. For the training configuration, we set the global batch size to 64, the learning rate to $2\times 10^{-7}$, the number of rollouts to 8, the rollout temperature to 0.7, and the KL coefficient to $5\times 10^{-3}$. The training was conducted for 2 epochs, consuming a total of approximately 770 GPU hours (equivalent NVIDIA A100 40G GPU hours).

\textbf{Hyperparameters.} In Stage 1, the reward weights were set as follows: $w_{\text{trans}}=3$ and $w_{\text{post}}=1.5$. In Stage 2, we adjusted these weights to $w_{\text{trans}}=5$ and $w_{\text{post}}=0.5$. These values are selected based on the ablation in Figure~\ref{fig:hp_analysis}(a) and Figure~\ref{fig:hp_analysis}(b), where each configuration achieves the best performance in its respective stage.

\textbf{Evaluation.} To evaluate the baseline models, we obtained open-source models from Hugging Face and accessed closed-source models via their official APIs. We utilized the official implementations and model checkpoints to compute the COMETKIWI-XL and MetricX metrics. Similarly, the GRF scores were obtained via API calls.

\begin{table*}[t]
\centering

\caption{Experimental results in Chinese-to-English translation. \textbf{Bold} and \underline{underlined} values denote the best and second-best scores, respectively.}

\resizebox{\textwidth}{!}
{
\begin{tabular}{lcccccccccccccccc}
\toprule[1pt]
\multirow{2}{*}{Model} & \multicolumn{4}{c}{Our} & \multicolumn{4}{c}{WMT23} & \multicolumn{4}{c}{WMT24} & \multicolumn{4}{c}{FLORES-200} \\
\cmidrule(lr){2-5} \cmidrule(lr){6-9} \cmidrule(lr){10-13} \cmidrule(lr){14-17}
& GRF & MX24 & CK & Token & GRF & MX24 & CK & Token & GRF & MX24 & CK & Token & GRF & MX24 & CK & Token \\
\midrule[1pt]
Qwen-7B-Instruct & 73.16 & 2.20 & 78.36 & \underline{21.31} & 82.00 & 2.54 & 78.11 & \underline{34.40} & 83.83 & 2.49 & 80.05 & 44.32 & 84.38 & 1.59 & 80.61 & 28.64 \\
Qwen2.5-14B & 79.89 & 2.30 & 79.97 & 26.00 & 87.27 & 2.46 & 80.46 & 46.32 & 86.35 & 2.58 & 81.66 & 55.92 & 88.98 & 1.83 & 85.24 & 32.45 \\
Qwen2.5-14B(w/ refl) & 86.91 & 1.67 & 80.78 & 183.72 & 90.65 & 1.72 & 81.15 & 210.63 & 90.09 & 2.15 & 81.96 & 225.70 & 92.91 & 0.99 & 85.77 & 641.89 \\
Llama3.1-8B & 75.57 & 2.25 & 79.98 & 25.87 & 79.86 & 2.60 & 79.56 & 37.53 & 81.77 & 2.68 & 80.44 & 57.79 & 84.12 & 1.81 & 83.53 & 27.23 \\
\midrule[1pt]
Qwen3-8B & 88.24 & 1.59 & 80.93 & 566.54 & 90.23 & 1.64 & 80.67 & 561.76 & 89.88 & 1.98 & 81.95 & 593.48 & 93.80 & 0.98 & 84.29 & 580.23 \\
Qwen3-8B(w/ refl) & 89.97 & 1.57 & 81.29 & 669.21 & \underline{92.03} & 1.63 & 81.38 & 698.51 & 90.77 & \underline{1.97} & \textbf{82.41} & 758.57 & \underline{95.14} & 0.98 & 85.77 & 681.00 \\
Marco-o1 & 80.24 & 1.81 & 80.35 & 490.37 & 87.70 & 2.01 & 80.34 & 498.71 & 86.15 & 2.28 & 80.56 & 495.59 & 90.20 & 1.30 & 85.31 & 477.49 \\
Marco-o1(w/ refl) & 86.08 & 1.68 & 80.44 & 602.75 & 88.54 & 1.77 & 80.54 & 574.07 & 87.65 & 2.09 & 81.09 & 650.91 & 93.03 & 1.05 & 85.33 & 584.38 \\
DeepSeek-R1 & 87.29 & 1.62 & 80.67 & 411.58 & 91.07 & 1.54 & 80.71 & 431.80 & 90.36 & \underline{1.97} & 81.91 & 542.17 & 93.32 & \underline{0.94} & \underline{85.86} & 526.94 \\
DeepSeek-R1(w/ refl) & \underline{90.02} & 1.59 & \underline{81.35} & 1567.80 & \textbf{92.26} & \underline{1.53} & \textbf{81.59} & 1631.93 & \textbf{91.39} & \textbf{1.94} & 82.09 & 1542.51 & 94.64 & \underline{0.94} & \textbf{85.97} & 1526.89 \\
QwQ-32B & 87.01 & 1.53 & 80.41 & 602.75 & 87.25 & 1.60 & 81.32 & 633.19 & 88.96 & 2.02 & 81.04 & 645.55 & 91.31 & 1.05 & 80.40 & 627.47 \\
QwQ-32B(w/ refl) & 88.67 & \underline{1.51} & 80.46 & 1000.16 & 89.09 & 1.58 & 81.38 & 1056.95 & \underline{91.34} & 2.02 & 81.78 & 1065.69 & 92.84 & 1.04 & 84.09 & 949.07 \\
\midrule
Qwen-7B-LoRA-v3(Cold Start) & 86.78 & 1.63 & 80.86 & \textbf{20.43} & 88.59 & 1.75 & 80.75 & \textbf{32.45} & 87.23 & 2.88 & 80.31 & \textbf{40.95} & 93.64 & 1.06 & 84.93 & \underline{27.22} \\
Qwen-7B-RL-v3(Our) & 88.88 & 1.58 & 81.22 & 23.71 & 89.94 & 1.59 & 81.27 & 34.91 & 89.71 & 2.51 & 81.98 & \underline{41.46} & 94.36 & 1.01 & 85.22 & \textbf{26.87} \\
Qwen-7B-RL-v3-Round2(Our) & \textbf{90.35} & \textbf{1.50} & \textbf{81.38} & 22.78 & 91.13 & \textbf{1.52} & \underline{81.49} & 34.50 & 90.07 & 2.11 & \underline{82.32} & 44.33 & \textbf{95.56} & \textbf{0.93} & 85.82 & 29.16 \\
\bottomrule[1pt]
\end{tabular}
}

\label{tab:zh-en}
\end{table*}

\section{Reflection Consolidation Prompt}
\label{app:Reflection Consolidation Prompt}
The following prompt consolidates multiple rounds of translation feedback into a single, coherent reflection that directly guides the revision from the initial translation to the final translation.

\begin{tcolorbox}[
    colback=gray!10!white,     
    colframe=black,            
    boxrule=1pt,               
    boxsep=4pt,                
    top=4pt, bottom=4pt,       
    left=8pt, right=8pt        
]
You are a translation revision editor. Your task is to consolidate multiple rounds of translation reflections into a unified set of revision feedback that can be used directly to revise the “Initial Translation” into the “Final Translation”.

\#\# Input

<initial\_translation>
{{initial\_translation}}
</initial\_translation>

<intermediate\_reflections>
{{intermediate\_reflections}}
</intermediate\_reflections>

<final\_translation>
{{final\_translation}}
</final\_translation>

\#\# Task Objective

Generate an “Aggregated Reflection” as the intermediate step in the following training sequence:

Initial Translation → Aggregated Reflection → Final Translation

The aggregated reflection must directly address the initial translation, identify specific issues that need revision, and provide revision directions consistent with the final translation. It should be standalone, coherent, and actionable revision feedback, rather than a summary of the multi-round reflections or a retrospective account of the revision process.

\#\# Consolidation Rules

1. Use the initial translation as the sole reference for locating issues
   - Every issue must correspond to something present in the initial translation.
   - Remove issues that apply only to intermediate versions and do not exist in the initial translation.
   - If an issue underwent multiple rounds of revision, consolidate them into a recommendation that leads directly from the initial translation to the final translation, without retaining intermediate attempts.

2. Use the final translation as the target version
   - Retain reflections that explain or guide the actual changes from the initial translation to the final translation.
   - Remove suggestions that were not adopted in the final translation, were overturned by subsequent reflections, or conflict with the final translation.
   - When suggestions from different rounds disagree, retain only the parts that are consistent with the final translation and supported by the input.
   - The final translation is the target version for this task, but this does not mean that every change necessarily corrects an objective error. Changes involving only wording or stylistic preferences should be described as refinements in expression, not overstated as corrections of mistranslations.

3. Merge duplicate content
   - If the same issue appears in multiple rounds of reflection, include it only once.
   - Combine complementary information into complete, concise recommendations.
   - Do not retain narratives of repeated experimentation, withdrawn suggestions, or the evolution of different versions.

4. Make recommendations specific and actionable
   - Each recommendation should, wherever possible, identify the specific location or wording in the initial translation, the aspect that needs adjustment, and a clear direction for revision.
   - Quote brief excerpts or provide local replacement wording when necessary to eliminate ambiguity.
   - Avoid vague comments such as “improve accuracy” or “make the wording more natural” without identifying a specific target.
   - If a change affects multiple locations, specify the scope of the consistent revision, such as terminology, forms of address, tense, or register.

5. Cover key changes without fabricating justifications
   - Prioritize key changes involving meaning, completeness of information, terminology, logical relationships, grammar, tone, and style.
   - If the multi-round reflections omit important changes between the initial and final translations, you may add revision recommendations directly supported by the two versions.
   - The input does not include the source text. Do not fabricate source content or present unsupported judgments as “The source text explicitly states...”
   - If you cannot determine whether a change corrects a mistranslation, describe only the observable adjustment in expression without inventing a semantic justification.
   - Do not force an explanation for differences that cannot be explained by the input.

\#\# Output Format

Output only the body of the aggregated reflection. Do not include task instructions, the analysis process, version comparison tables, or any other additional content.
\end{tcolorbox}

\section{Our Adapt thinking Baseline}
\label{app:Adapt}

Our pre-thinking baseline uses training data aligned with our method, employing DeepSeek-V3 to generate pre-thinking data for SFT cold start. The data format is as follows:

\begin{tcolorbox}[
    colback=gray!10!white,     
    colframe=black,            
    boxrule=1pt,               
    boxsep=4pt,                
    top=4pt, bottom=4pt,       
    left=8pt, right=8pt        
]
\begin{CJK}{UTF8}{gbsn}
<think>首先，"kind of"是一个常见的俚语表达，通常用来表示某种程度的不确定性或轻微的程度。然后，"Riverdale"是一个专有名词，但在这个上下文中，它可能指的是一个地方名，也可能是电视剧《Riverdale》中的Riverdale镇。如果是指地点，那么“his last meal in Riverdale”可以理解为他在Riverdale的最后一次用餐；如果是指电视剧，则可能是在描述剧中人物即将离开或结束在Riverdale的生活。考虑到俚语的使用，这句句子更有可能出自一种非正式的语境，比如电视剧的讨论或者个人的情感表达。因此，我将假设此句出自电视剧的讨论背景。</think><answer>有点希望这是他在Riverdale的最后一顿饭</answer>
\end{CJK}
\end{tcolorbox}
In datasets where no reflection is required, the `<think></think>` tag remains empty. The ratio of reflective to non-reflective data is set at 2:3. Using LLaMA-Factory for fine-tuning and verl for GRPO training, we cultivated a model capable of adaptively deciding whether to engage in reflection before generating a response. This adaptive reflection method allows the model to avoid excessive contemplation on simple tasks, thereby preventing performance degradation. Furthermore, it effectively reduces token consumption, enhancing overall efficiency.

\section{Multilingual}
\label{app:multilingual}

Across Chinese-to-English, English-to-Russian, and English-to-German translation, ReflectMT substantially reduces generated tokens compared with explicit-reflection baselines while achieving competitive translation quality. Detailed results for the three translation directions are reported in Tables~\ref{tab:zh-en}, \ref{tab:en-ru}, and~\ref{tab:en-de}, respectively.


\begin{table*}[t]
\centering

\caption{Experimental results in English-to-German translation. \textbf{Bold} and \underline{underlined} values denote the best and second-best scores, respectively.}

\resizebox{\textwidth}{!}
{
\begin{tabular}{lcccccccccccc}
\toprule[1pt]
\multirow{2}{*}{Model} & \multicolumn{4}{c}{WMT23} & \multicolumn{4}{c}{WMT24} & \multicolumn{4}{c}{FLORES-200} \\
\cmidrule(lr){2-5} \cmidrule(lr){6-9} \cmidrule(lr){10-13}
& GRF & MX24 & CK & Token & GRF & MX24 & CK & Token & GRF & MX24 & CK & Token \\
\midrule[1pt]
Qwen-7B-Instruct & 59.04 & 4.07 & 74.01 & 142.50 & 57.25 & 3.55 & 74.89 & \underline{63.34} & 54.10 & 2.36 & 79.21 & 42.36 \\
Qwen2.5-14B & 72.91 & 2.82 & 78.12 & 139.04 & 67.41 & 2.45 & 79.50 & 75.30 & 77.52 & 1.07 & 83.80 & 48.24 \\
Qwen2.5-14B(w/ refl) & 78.84 & 2.77 & 79.08 & 397.14 & 69.54 & 2.19 & 80.09 & 284.27 & 78.32 & 1.00 & 84.53 & 224.11 \\
Llama3.1-8B & 70.11 & 3.11 & 74.35 & \textbf{112.08} & 57.66 & 2.42 & 74.61 & 69.48 & 72.28 & 1.01 & 84.63 & \underline{40.18} \\
\midrule[1pt]
Qwen3-8B & 69.82 & 2.61 & 79.93 & 716.42 & 73.01 & 2.27 & 80.87 & 491.98 & 76.74 & 0.86 & 84.93 & 592.83 \\
Qwen3-8B(w/ refl) & 70.56 & 2.64 & 80.25 & 860.14 & 75.32 & 2.18 & \underline{80.91} & 655.36 & \underline{78.66} & 0.80 & 85.67 & 689.12 \\
Marco-o1 & 65.51 & 3.28 & 76.96 & 616.28 & 59.98 & 2.81 & 77.52 & 466.15 & 61.69 & 1.39 & 82.54 & 345.46 \\
Marco-o1(w/ refl) & 66.40 & 3.23 & 77.19 & 849.95 & 62.64 & 2.56 & 78.60 & 618.26 & 64.82 & 1.26 & 83.21 & 639.70 \\
DeepSeek-R1 & 79.42 & 2.15 & 81.66 & 820.33 & 75.80 & 1.73 & 80.83 & 690.31 & 73.33 & \underline{0.61} & \underline{86.71} & 840.02 \\
DeepSeek-R1(w/ refl) & \underline{80.13} & \underline{2.05} & \underline{81.90} & 1360.79 & \textbf{78.66} & \underline{1.69} & \textbf{81.77} & 1060.92 & \textbf{79.32} & \textbf{0.60} & \textbf{86.75} & 1139.90 \\
\midrule
Qwen-7B-LoRA-v3(Cold Start) & 73.67 & 3.71 & 74.98 & 124.09 & 62.04 & 3.25 & 75.19 & 65.67 & 70.23 & 1.48 & 81.36 & \textbf{40.10} \\
Qwen-7B-RL-v3(Our) & 75.39 & 2.63 & 78.37 & 130.92 & 64.15 & 2.49 & 77.53 & 64.28 & 72.03 & 1.14 & 83.97 & 43.82 \\
Qwen-7B-RL-v3-Round2(Our) & \textbf{80.83} & \textbf{2.00} & \textbf{82.51} & \underline{119.14} & \underline{77.10} & \textbf{1.38} & 80.82 & \textbf{59.49} & 78.46 & 0.68 & 86.22 & 40.56 \\
\bottomrule[1pt]
\end{tabular}
}

\label{tab:en-de}
\end{table*}

\begin{table*}[t]
\centering

\caption{Experimental results in English-to-Russian translation. \textbf{Bold} and \underline{underlined} values denote the best and second-best scores, respectively.}

\resizebox{\textwidth}{!}
{
\begin{tabular}{lcccccccccccc}
\toprule[1pt]
\multirow{2}{*}{Model} & \multicolumn{4}{c}{WMT23} & \multicolumn{4}{c}{WMT24} & \multicolumn{4}{c}{FLORES-200} \\
\cmidrule(lr){2-5} \cmidrule(lr){6-9} \cmidrule(lr){10-13}
& GRF & MX24 & CK & Token & GRF & MX24 & CK & Token & GRF & MX24 & CK & Token \\
\midrule[1pt]
Qwen-7B-Instruct & 63.98 & 3.90 & 77.69 & \underline{52.34} & 56.07 & 5.50 & 73.24 & 94.69 & 68.49 & 3.81 & 79.13 & 65.93 \\
Qwen2.5-14B & 72.61 & 3.61 & 81.31 & 72.93 & 61.72 & 4.07 & 78.01 & 91.92 & 74.27 & 2.05 & 83.96 & 91.36 \\
Qwen2.5-14B(w/ refl) & 76.00 & 2.80 & 74.43 & 268.44 & 65.88 & 3.84 & 79.18 & 276.94 & 82.92 & 1.95 & 85.20 & 274.99 \\
Llama3.1-8B & 70.74 & 3.87 & 78.87 & 64.42 & 50.98 & 4.83 & 74.31 & \underline{66.77} & 74.14 & 2.27 & 85.09 & \underline{45.80} \\
\midrule[1pt]
Qwen3-8B & 74.72 & 3.10 & 81.67 & 487.80 & 72.59 & 4.14 & 79.60 & 683.68 & 74.20 & 2.91 & 85.38 & 524.97 \\
Qwen3-8B(w/ refl) & 75.09 & 2.85 & 82.27 & 685.75 & 72.95 & 4.11 & 79.66 & 761.59 & 75.09 & 2.80 & 86.06 & 634.82 \\
Marco-o1 & 68.56 & 3.38 & 80.29 & 450.95 & 61.64 & 4.68 & 76.58 & 591.11 & 68.29 & 2.57 & 83.74 & 550.43 \\
Marco-o1(w/ refl) & 69.20 & 3.48 & 79.40 & 549.94 & 68.34 & 4.56 & 77.48 & 768.73 & 73.54 & 2.46 & 84.65 & 601.34 \\
DeepSeek-R1 & 81.94 & \underline{2.36} & 83.38 & 748.59 & \underline{81.28} & \underline{3.16} & \underline{80.42} & 985.93 & 79.82 & 1.33 & 87.69 & 974.30 \\
DeepSeek-R1(w/ refl) & \underline{83.07} & \textbf{2.12} & \textbf{84.06} & 967.00 & \textbf{81.50} & \textbf{3.15} & \textbf{81.00} & 1138.65 & 83.75 & \underline{1.28} & \textbf{87.94} & 1226.05 \\
\midrule
Qwen-7B-LoRA-v3(Cold Start) & 73.20 & 3.81 & 81.58 & \textbf{37.89} & 71.56 & 4.29 & 78.08 & \textbf{64.74} & 77.10 & 2.09 & 85.74 & \textbf{44.19} \\
Qwen-7B-RL-v3(Our) & 77.37 & 2.71 & 83.14 & 58.83 & 75.65 & 3.55 & 79.76 & 74.29 & \underline{86.76} & 1.30 & 85.81 & 49.24 \\
Qwen-7B-RL-v3-Round2(Our) & \textbf{83.75} & \textbf{2.12} & \underline{83.98} & 61.78 & 78.38 & 3.37 & 80.23 & 87.83 & \textbf{87.01} & \textbf{1.26} & \underline{87.77} & 68.76 \\
\bottomrule[1pt]
\end{tabular}
}

\label{tab:en-ru}
\end{table*}

\section{Cultural Adaptation in Reflection}
\label{app:cultural_adaptation}

The reflector does not follow a fixed rule of preserving, omitting, or replacing culturally specific expressions. Our qualitative inspection of culturally marked examples sampled from SlangDIT and MetaphorTrans suggests that it generally identifies the expression's pragmatic function and favors a natural target-language paraphrase or functional equivalent. Source-culture references may be retained when understandable, while the intended strategy avoids omitting source meaning or introducing unsupported target-culture content.

\paragraph{Case Study: ``Wigs on the green.''}
\begin{CJK}{UTF8}{gbsn}
This British idiom refers to an intense dispute or chaotic confrontation. In the inspected example, the reflector judges the initial Chinese rendering ``大闹'' to be overly colloquial and insufficiently expressive of the intended implication, and subsequently revises it to ``陷入一片混乱之中.'' The translation quality score increases from 65 to 85. Rather than reproducing the opaque surface form, the revision conveys the expression through a natural target-language paraphrase. 

\end{CJK}

\section{GPT-4o Score}
\label{app:GPT}

\begin{table}[t]
\centering
\caption{Performance metrics for GPT-4o and ReflectMT-Stage2.}

\resizebox{0.48\textwidth}{!}
{
\begin{tabular}{lcccc}
\toprule
\textbf{Model} & \textbf{GRF} & \textbf{MX24} & \textbf{CK} & \textbf{Token} \\
\midrule
GPT-4o & 77.62 & 2.63 & 77.97 & 28.57 \\
ReflectMT-Stage2 (Our) & 82.55 & 2.00 & 78.05 & 30.73 \\
\bottomrule
\end{tabular}
}
\label{tab:performance}
\end{table}

Due to the costs associated with using the GPT-4o API, we limited our testing of GPT-4o to our custom dataset. As shown in Table \ref{tab:performance}, ReflectMT achieves higher GRF scores compared to GPT-4o, while maintaining similar token consumption.


\newpage
\section*{NeurIPS Paper Checklist}

\begin{enumerate}

\item {\bf Claims}
    \item[] Question: Do the main claims made in the abstract and introduction accurately reflect the paper's contributions and scope?
    \item[] Answer: \answerYes{} 
    \item[] Justification: The claims made in the abstract and introduction are empirically verified in Section \ref{exper}, in particular Table~\ref{fig:main}
    \item[] Guidelines:
    \begin{itemize}
        \item The answer \answerNA{} means that the abstract and introduction do not include the claims made in the paper.
        \item The abstract and/or introduction should clearly state the claims made, including the contributions made in the paper and important assumptions and limitations. A \answerNo{} or \answerNA{} answer to this question will not be perceived well by the reviewers. 
        \item The claims made should match theoretical and experimental results, and reflect how much the results can be expected to generalize to other settings. 
        \item It is fine to include aspirational goals as motivation as long as it is clear that these goals are not attained by the paper. 
    \end{itemize}

\item {\bf Limitations}
    \item[] Question: Does the paper discuss the limitations of the work performed by the authors?
    \item[] Answer: \answerYes{} 
    \item[] Justification: We mention our limitations in Section \ref{limit}.
    \item[] Guidelines:
    \begin{itemize}
        \item The answer \answerNA{} means that the paper has no limitation while the answer \answerNo{} means that the paper has limitations, but those are not discussed in the paper. 
        \item The authors are encouraged to create a separate ``Limitations'' section in their paper.
        \item The paper should point out any strong assumptions and how robust the results are to violations of these assumptions (e.g., independence assumptions, noiseless settings, model well-specification, asymptotic approximations only holding locally). The authors should reflect on how these assumptions might be violated in practice and what the implications would be.
        \item The authors should reflect on the scope of the claims made, e.g., if the approach was only tested on a few datasets or with a few runs. In general, empirical results often depend on implicit assumptions, which should be articulated.
        \item The authors should reflect on the factors that influence the performance of the approach. For example, a facial recognition algorithm may perform poorly when image resolution is low or images are taken in low lighting. Or a speech-to-text system might not be used reliably to provide closed captions for online lectures because it fails to handle technical jargon.
        \item The authors should discuss the computational efficiency of the proposed algorithms and how they scale with dataset size.
        \item If applicable, the authors should discuss possible limitations of their approach to address problems of privacy and fairness.
        \item While the authors might fear that complete honesty about limitations might be used by reviewers as grounds for rejection, a worse outcome might be that reviewers discover limitations that aren't acknowledged in the paper. The authors should use their best judgment and recognize that individual actions in favor of transparency play an important role in developing norms that preserve the integrity of the community. Reviewers will be specifically instructed to not penalize honesty concerning limitations.
    \end{itemize}

\item {\bf Theory assumptions and proofs}
    \item[] Question: For each theoretical result, does the paper provide the full set of assumptions and a complete (and correct) proof?
    \item[] Answer: \answerNA{} 
    \item[] Justification: This paper does not include any theoretical results or formal proofs; our contribution is empirical, focusing on a training framework and experimental validation.
    \item[] Guidelines:
    \begin{itemize}
        \item The answer \answerNA{} means that the paper does not include theoretical results. 
        \item All the theorems, formulas, and proofs in the paper should be numbered and cross-referenced.
        \item All assumptions should be clearly stated or referenced in the statement of any theorems.
        \item The proofs can either appear in the main paper or the supplemental material, but if they appear in the supplemental material, the authors are encouraged to provide a short proof sketch to provide intuition. 
        \item Inversely, any informal proof provided in the core of the paper should be complemented by formal proofs provided in appendix or supplemental material.
        \item Theorems and Lemmas that the proof relies upon should be properly referenced. 
    \end{itemize}

    \item {\bf Experimental result reproducibility}
    \item[] Question: Does the paper fully disclose all the information needed to reproduce the main experimental results of the paper to the extent that it affects the main claims and/or conclusions of the paper (regardless of whether the code and data are provided or not)?
    \item[] Answer: \answerYes{} 
    \item[] Justification: We describe the full methodology in Section 3, the data construction pipeline in Section 2, and provide detailed hyperparameters, training configurations, backbones, and evaluation prompts in Appendices A–D (e.g., learning rate, batch size, GRPO settings, reward weights, temperature).

    \item[] Guidelines:
    \begin{itemize}
        \item The answer \answerNA{} means that the paper does not include experiments.
        \item If the paper includes experiments, a \answerNo{} answer to this question will not be perceived well by the reviewers: Making the paper reproducible is important, regardless of whether the code and data are provided or not.
        \item If the contribution is a dataset and\slash or model, the authors should describe the steps taken to make their results reproducible or verifiable. 
        \item Depending on the contribution, reproducibility can be accomplished in various ways. For example, if the contribution is a novel architecture, describing the architecture fully might suffice, or if the contribution is a specific model and empirical evaluation, it may be necessary to either make it possible for others to replicate the model with the same dataset, or provide access to the model. In general. releasing code and data is often one good way to accomplish this, but reproducibility can also be provided via detailed instructions for how to replicate the results, access to a hosted model (e.g., in the case of a large language model), releasing of a model checkpoint, or other means that are appropriate to the research performed.
        \item While NeurIPS does not require releasing code, the conference does require all submissions to provide some reasonable avenue for reproducibility, which may depend on the nature of the contribution. For example
        \begin{enumerate}
            \item If the contribution is primarily a new algorithm, the paper should make it clear how to reproduce that algorithm.
            \item If the contribution is primarily a new model architecture, the paper should describe the architecture clearly and fully.
            \item If the contribution is a new model (e.g., a large language model), then there should either be a way to access this model for reproducing the results or a way to reproduce the model (e.g., with an open-source dataset or instructions for how to construct the dataset).
            \item We recognize that reproducibility may be tricky in some cases, in which case authors are welcome to describe the particular way they provide for reproducibility. In the case of closed-source models, it may be that access to the model is limited in some way (e.g., to registered users), but it should be possible for other researchers to have some path to reproducing or verifying the results.
        \end{enumerate}
    \end{itemize}

\item {\bf Open access to data and code}
    \item[] Question: Does the paper provide open access to the data and code, with sufficient instructions to faithfully reproduce the main experimental results, as described in supplemental material?
    \item[] Answer: \answerYes{} 
    \item[] Justification: Code, dataset, training scripts and prompts are publicly released upon publication to ensure full reproducibility.

    \item[] Guidelines:
    \begin{itemize}
        \item The answer \answerNA{} means that paper does not include experiments requiring code.
        \item Please see the NeurIPS code and data submission guidelines (\url{https://neurips.cc/public/guides/CodeSubmissionPolicy}) for more details.
        \item While we encourage the release of code and data, we understand that this might not be possible, so \answerNo{} is an acceptable answer. Papers cannot be rejected simply for not including code, unless this is central to the contribution (e.g., for a new open-source benchmark).
        \item The instructions should contain the exact command and environment needed to run to reproduce the results. See the NeurIPS code and data submission guidelines (\url{https://neurips.cc/public/guides/CodeSubmissionPolicy}) for more details.
        \item The authors should provide instructions on data access and preparation, including how to access the raw data, preprocessed data, intermediate data, and generated data, etc.
        \item The authors should provide scripts to reproduce all experimental results for the new proposed method and baselines. If only a subset of experiments are reproducible, they should state which ones are omitted from the script and why.
        \item At submission time, to preserve anonymity, the authors should release anonymized versions (if applicable).
        \item Providing as much information as possible in supplemental material (appended to the paper) is recommended, but including URLs to data and code is permitted.
    \end{itemize}

\item {\bf Experimental setting/details}
    \item[] Question: Does the paper specify all the training and test details (e.g., data splits, hyperparameters, how they were chosen, type of optimizer) necessary to understand the results?
    \item[] Answer: \answerYes{} 
    \item[] Justification: Section \ref{subsec:4.1} describes the datasets, metrics, and backbones. Appendix \ref{app:Details} provides full implementation details, including SFT hyperparameters, GRPO settings , reward weights for both stages, and evaluation configurations. Dataset splits are given in Table \ref{tab:dataset}.

    \item[] Guidelines:
    \begin{itemize}
        \item The answer \answerNA{} means that the paper does not include experiments.
        \item The experimental setting should be presented in the core of the paper to a level of detail that is necessary to appreciate the results and make sense of them.
        \item The full details can be provided either with the code, in appendix, or as supplemental material.
    \end{itemize}

\item {\bf Experiment statistical significance}
    \item[] Question: Does the paper report error bars suitably and correctly defined or other appropriate information about the statistical significance of the experiments?
    \item[] Answer: \answerYes{} 
    \item[] Justification: We conduct paired bootstrap significance tests, report multi‑seed RL replicates in Appendix \ref{app:Robustness Analysis} to assess result stability. We do not plot per‑run error bars in main text figures due to high computational cost of full end‑to‑end RL training. See Appendix \ref{app:Robustness Analysis} for bootstrap p‑values and multi‑seed results.

    \item[] Guidelines:
    \begin{itemize}
        \item The answer \answerNA{} means that the paper does not include experiments.
        \item The authors should answer \answerYes{} if the results are accompanied by error bars, confidence intervals, or statistical significance tests, at least for the experiments that support the main claims of the paper.
        \item The factors of variability that the error bars are capturing should be clearly stated (for example, train/test split, initialization, random drawing of some parameter, or overall run with given experimental conditions).
        \item The method for calculating the error bars should be explained (closed form formula, call to a library function, bootstrap, etc.)
        \item The assumptions made should be given (e.g., Normally distributed errors).
        \item It should be clear whether the error bar is the standard deviation or the standard error of the mean.
        \item It is OK to report 1-sigma error bars, but one should state it. The authors should preferably report a 2-sigma error bar than state that they have a 96\% CI, if the hypothesis of Normality of errors is not verified.
        \item For asymmetric distributions, the authors should be careful not to show in tables or figures symmetric error bars that would yield results that are out of range (e.g., negative error rates).
        \item If error bars are reported in tables or plots, the authors should explain in the text how they were calculated and reference the corresponding figures or tables in the text.
    \end{itemize}

\item {\bf Experiments compute resources}
    \item[] Question: For each experiment, does the paper provide sufficient information on the computer resources (type of compute workers, memory, time of execution) needed to reproduce the experiments?
    \item[] Answer: \answerYes{} 
    \item[] Justification: Appendix~\ref{app:Details} reports the compute budget in terms of NVIDIA A100 40GB-equivalent GPU hours. Due to institutional policy, we do not disclose the specific GPU model used in our internal cluster. DeepSeek-V3 (671B) is locally deployed and serves as both the data-construction agent and the LLM-as-a-judge reward model.
    \item[] Guidelines:
    \begin{itemize}
        \item The answer \answerNA{} means that the paper does not include experiments.
        \item The paper should indicate the type of compute workers CPU or GPU, internal cluster, or cloud provider, including relevant memory and storage.
        \item The paper should provide the amount of compute required for each of the individual experimental runs as well as estimate the total compute. 
        \item The paper should disclose whether the full research project required more compute than the experiments reported in the paper (e.g., preliminary or failed experiments that didn't make it into the paper). 
    \end{itemize}
    
\item {\bf Code of ethics}
    \item[] Question: Does the research conducted in the paper conform, in every respect, with the NeurIPS Code of Ethics \url{https://neurips.cc/public/EthicsGuidelines}?
    \item[] Answer: \answerYes{} 
    \item[] Justification: We have reviewed the NeurIPS Code of Ethics and confirm that our research conforms to it. Our work uses publicly available datasets and does not involve human subjects or sensitive data.

    \item[] Guidelines:
    \begin{itemize}
        \item The answer \answerNA{} means that the authors have not reviewed the NeurIPS Code of Ethics.
        \item If the authors answer \answerNo, they should explain the special circumstances that require a deviation from the Code of Ethics.
        \item The authors should make sure to preserve anonymity (e.g., if there is a special consideration due to laws or regulations in their jurisdiction).
    \end{itemize}

\item {\bf Broader impacts}
    \item[] Question: Does the paper discuss both potential positive societal impacts and negative societal impacts of the work performed?
    \item[] Answer: \answerNA{} 
    \item[] Justification: This work focuses on improving the efficiency and quality of machine translation, a well-established and widely-used technology. It does not introduce new capabilities with direct foreseeable negative societal impacts beyond those generally associated with MT systems (e.g., occasional mistranslation).

    \item[] Guidelines:
    \begin{itemize}
        \item The answer \answerNA{} means that there is no societal impact of the work performed.
        \item If the authors answer \answerNA{} or \answerNo, they should explain why their work has no societal impact or why the paper does not address societal impact.
        \item Examples of negative societal impacts include potential malicious or unintended uses (e.g., disinformation, generating fake profiles, surveillance), fairness considerations (e.g., deployment of technologies that could make decisions that unfairly impact specific groups), privacy considerations, and security considerations.
        \item The conference expects that many papers will be foundational research and not tied to particular applications, let alone deployments. However, if there is a direct path to any negative applications, the authors should point it out. For example, it is legitimate to point out that an improvement in the quality of generative models could be used to generate Deepfakes for disinformation. On the other hand, it is not needed to point out that a generic algorithm for optimizing neural networks could enable people to train models that generate Deepfakes faster.
        \item The authors should consider possible harms that could arise when the technology is being used as intended and functioning correctly, harms that could arise when the technology is being used as intended but gives incorrect results, and harms following from (intentional or unintentional) misuse of the technology.
        \item If there are negative societal impacts, the authors could also discuss possible mitigation strategies (e.g., gated release of models, providing defenses in addition to attacks, mechanisms for monitoring misuse, mechanisms to monitor how a system learns from feedback over time, improving the efficiency and accessibility of ML).
    \end{itemize}
    
\item {\bf Safeguards}
    \item[] Question: Does the paper describe safeguards that have been put in place for responsible release of data or models that have a high risk for misuse (e.g., pre-trained language models, image generators, or scraped datasets)?
    \item[] Answer: \answerNA{} 
    \item[] Justification: Our released model and dataset target standard English-to-Chinese machine translation and pose no higher risk of misuse than existing publicly available MT models and corpora. Therefore no additional safeguards are required.

    \item[] Guidelines:
    \begin{itemize}
        \item The answer \answerNA{} means that the paper poses no such risks.
        \item Released models that have a high risk for misuse or dual-use should be released with necessary safeguards to allow for controlled use of the model, for example by requiring that users adhere to usage guidelines or restrictions to access the model or implementing safety filters. 
        \item Datasets that have been scraped from the Internet could pose safety risks. The authors should describe how they avoided releasing unsafe images.
        \item We recognize that providing effective safeguards is challenging, and many papers do not require this, but we encourage authors to take this into account and make a best faith effort.
    \end{itemize}

\item {\bf Licenses for existing assets}
    \item[] Question: Are the creators or original owners of assets (e.g., code, data, models), used in the paper, properly credited and are the license and terms of use explicitly mentioned and properly respected?
    \item[] Answer: \answerYes{} 
    \item[] Justification: All used assets are properly cited, including base models, datasets, and frameworks. We use them in accordance with their licenses and terms of use.

    \item[] Guidelines:
    \begin{itemize}
        \item The answer \answerNA{} means that the paper does not use existing assets.
        \item The authors should cite the original paper that produced the code package or dataset.
        \item The authors should state which version of the asset is used and, if possible, include a URL.
        \item The name of the license (e.g., CC-BY 4.0) should be included for each asset.
        \item For scraped data from a particular source (e.g., website), the copyright and terms of service of that source should be provided.
        \item If assets are released, the license, copyright information, and terms of use in the package should be provided. For popular datasets, \url{paperswithcode.com/datasets} has curated licenses for some datasets. Their licensing guide can help determine the license of a dataset.
        \item For existing datasets that are re-packaged, both the original license and the license of the derived asset (if it has changed) should be provided.
        \item If this information is not available online, the authors are encouraged to reach out to the asset's creators.
    \end{itemize}

\item {\bf New assets}
    \item[] Question: Are new assets introduced in the paper well documented and is the documentation provided alongside the assets?
    \item[] Answer: \answerYes{} 
    \item[] Justification: We introduce a new reflective English-to-Chinese translation dataset. Its construction pipeline, statistics, filtering rules, hyperparameters, and representative examples are fully documented in Section 2 and Appendix A. The dataset will be released with documentation upon acceptance.

    \item[] Guidelines:
    \begin{itemize}
        \item The answer \answerNA{} means that the paper does not release new assets.
        \item Researchers should communicate the details of the dataset\slash code\slash model as part of their submissions via structured templates. This includes details about training, license, limitations, etc. 
        \item The paper should discuss whether and how consent was obtained from people whose asset is used.
        \item At submission time, remember to anonymize your assets (if applicable). You can either create an anonymized URL or include an anonymized zip file.
    \end{itemize}

\item {\bf Crowdsourcing and research with human subjects}
    \item[] Question: For crowdsourcing experiments and research with human subjects, does the paper include the full text of instructions given to participants and screenshots, if applicable, as well as details about compensation (if any)? 
    \item[] Answer: \answerNA{} 
    \item[] Justification: This work does not use crowdsourcing or recruit external human‑subject participants. Blinded human evaluation reported in Appendix B is completed by internal researchers of our research group who are not listed as paper authors; no paid external annotators are involved.

    \item[] Guidelines:
    \begin{itemize}
        \item The answer \answerNA{} means that the paper does not involve crowdsourcing nor research with human subjects.
        \item Including this information in the supplemental material is fine, but if the main contribution of the paper involves human subjects, then as much detail as possible should be included in the main paper. 
        \item According to the NeurIPS Code of Ethics, workers involved in data collection, curation, or other labor should be paid at least the minimum wage in the country of the data collector. 
    \end{itemize}

\item {\bf Institutional review board (IRB) approvals or equivalent for research with human subjects}
    \item[] Question: Does the paper describe potential risks incurred by study participants, whether such risks were disclosed to the subjects, and whether Institutional Review Board (IRB) approvals (or an equivalent approval/review based on the requirements of your country or institution) were obtained?
    \item[] Answer: \answerNA{} 
    \item[] Justification: This paper does not involve research with human subjects; therefore no IRB approval is required.

    \item[] Guidelines:
    \begin{itemize}
        \item The answer \answerNA{} means that the paper does not involve crowdsourcing nor research with human subjects.
        \item Depending on the country in which research is conducted, IRB approval (or equivalent) may be required for any human subjects research. If you obtained IRB approval, you should clearly state this in the paper. 
        \item We recognize that the procedures for this may vary significantly between institutions and locations, and we expect authors to adhere to the NeurIPS Code of Ethics and the guidelines for their institution. 
        \item For initial submissions, do not include any information that would break anonymity (if applicable), such as the institution conducting the review.
    \end{itemize}

\item {\bf Declaration of LLM usage}
    \item[] Question: Does the paper describe the usage of LLMs if it is an important, original, or non-standard component of the core methods in this research? Note that if the LLM is used only for writing, editing, or formatting purposes and does \emph{not} impact the core methodology, scientific rigor, or originality of the research, declaration is not required.
    \item[] Answer: \answerYes{} 
    \item[] Justification: LLMs are a core component of our methodology. We use DeepSeek-V3 for reflective data construction (as Translator and Reflector agents) and as the LLM-as-a-Judge reward model during RL training. Qwen2.5-7B-Instruct serves as our base policy model. All usages are described in detail in Sections \ref{subsec:2.2} and \ref{method} and Appendix \ref{app:Evaluation Prompt}.

    \item[] Guidelines:
    \begin{itemize}
        \item The answer \answerNA{} means that the core method development in this research does not involve LLMs as any important, original, or non-standard components.
        \item Please refer to our LLM policy in the NeurIPS handbook for what should or should not be described.
    \end{itemize}

\end{enumerate}

\end{document}